\documentclass[1p,11pt,authoryear]{elsarticle}

\makeatletter
\def\ps@pprintTitle{%
\let\@oddhead\@empty
\let\@evenhead\@empty
\def\@oddfoot{\footnotesize\itshape
Preprint accepted for publication in \@journal, Elsevier.\hfill}%
\let\@evenfoot\@oddfoot}
\makeatother

\usepackage{hyperref}
\usepackage{url}

\usepackage{cleveref}

\usepackage[ruled]{algorithm2e}
\usepackage{amsmath}
\usepackage{amssymb}
\usepackage{bm}

\usepackage{graphicx}
\usepackage{comment}
\usepackage{siunitx}
\usepackage{wrapfig}
\usepackage{todonotes}
\usepackage{enumitem}
\usepackage{siunitx}
\usepackage{dsfont}
\usepackage{bbm}

\usepackage{breqn}
\usepackage{changepage}
\usepackage{natbib}

\newcommand{\real}{\mathbb{R}}

\newcommand{\be}{\begin{equation}}
\newcommand{\ee}{\end{equation}}
\newcommand{\bea}{\begin{eqnarray}}
\newcommand{\eea}{\end{eqnarray}}
\newcommand{\beaa}{\begin{eqnarray*}}
\newcommand{\eeaa}{\end{eqnarray*}}

\DeclareMathAlphabet{\mathpzc}{OT1}{pzc}{m}{n}

\DeclareMathOperator*{\argmax}{arg\,max}

\journal{Expert Systems with Applications}

\begin{document}

\begin{frontmatter}

\author[NDG]{Nikolaos D. Goumagias\corref{cor1}}
\ead{nikolaos.goumagias@northumbria.ac.uk}
\cortext[cor1]{Corresponding author}
\address[NDG]{Northumbria University, Newcastle Business School, Central Campus East 1, Newcastle upon Tyne, NE1 8ST, UK}
\author[DCV]{Dimitrios Hristu-Varsakelis}
\ead{dcv@uom.gr}
\address[DCV]{University of Macedonia, Department of Applied Informatics, Egnatia 156, Thessaloniki, 54006, Greece}
\author[YA]{Yannis M. Assael}
\ead{yannis.assael@cs.ox.ac.uk}
\address[YA]{University of Oxford, Department of Computer Science, Wolfson Building, Parks Road, Oxford, OX1 3QD, UK 
\vspace*{-1cm}}

\vspace*{-1cm}
\title{Using deep Q-learning to understand the tax evasion behavior of risk-averse firms}

\address{}
\vspace*{-1cm}
\begin{abstract}
Designing tax policies that are effective in curbing tax evasion and maximize state revenues requires a rigorous understanding of taxpayer behavior. This work explores the problem of determining the strategy a self-interested, risk-averse tax entity is expected to follow,
as it ``navigates'' - in the context of a Markov Decision Process -  a government-controlled tax environment that includes random audits, penalties and occasional tax amnesties. Although simplified versions of this problem have been previously explored, the mere assumption of risk-aversion (as opposed to risk-neutrality) raises the complexity of finding the optimal policy well beyond the reach of analytical techniques. Here, we obtain approximate solutions via a combination of Q-learning and recent advances in Deep Reinforcement Learning. 
By doing so, we i) determine the tax evasion behavior expected of the taxpayer entity, ii) calculate the degree of risk aversion of the ``average'' entity given empirical estimates of tax evasion, and iii) evaluate sample tax policies, in terms of expected revenues. Our model can be useful as a testbed for ``in-vitro'' testing of tax policies, while our results lead to various policy recommendations.
 \end{abstract}

\end{frontmatter}

\newpage
\section{Introduction}
\label{sec:introduction}
While the aftershocks of the latest global financial crisis are still being felt, many governments struggle to implement  public policy because of budget deficits or lagging tax revenues \citep{bayer14}. The latter problem arises as a result of reduced economic activity, or when there is a strong sense among taxpayers that the expected personal  benefit from tax evasion surpasses the corresponding social benefit of paying taxes \citep{Alm90,bornstein1990role}. This, in the absence of a properly designed tax system and enforcement mechanism, leads to tax evasion, a serious crime that saps the State of revenue and undermines the sense of social justice, as dishonest taxpayers seem to enjoy the same public goods as honest ones do. The resulting ``shadow economy'' also has a strong adverse impact on credit ratings and lending costs \citep{markellos2016sovereign}, welfare programs, fiscal policies and unemployment \citep{fleming2000shadow}. 

Of course, tax systems typically contain various safeguards to discourage tax evasion (defined here as the deliberate failure to declare all or some of one’s income to tax authorities). In practice, however, tax systems are rather complex policy structures which are difficult to make ``air-tight'' in terms of tax evasion, for reasons having to do with i) occasional ambiguity in tax regulations, which hinders tax compliance and enforcement \citep{andreoni1998tax}, and ii) the heterogeneous behaviors of the various taxpayer entities, based on their individual risk preferences \citep{hokamp2010income}. 

This paper is concerned with the development of a rigorous computational framework which 
can describe and predict the behavior of tax evaders, assuming that they are self-interested and work to maximize the utility of their own revenues, balancing potential gains from tax evasion against the risk of getting caught. In particular, we are interested in i) estimating State revenues for any given set of tax parameters (e.g., tax rates and penalties), ii) testing whether specific tax regulations are helpful or not, and iii) predicting how taxpayers - and tax compliance -  respond to parameter changes. This last item is linked to taxpayers' risk aversion, knowledge of which would help the State determine the effects of, for example, an increase in tax penalties or audit rates. 

The issues raised above are essential to the State if it is to know the extent to which its tax policies are working or to rank alternative policies and take steps towards maximizing revenue. In this work, we propose to explore them using a combination of deep neural networks and Q-learning for determining the tax evasion behavior of a risk averse taxpayer (we will use the term ``firm'' henceforth because we will be interested mainly in business entities). We will  develop and test our approach in a context that builds on the work of \citet{goumagias2012decision} (where only the case of risk neutrality was analyzed) and involves a close-to-real-world tax system, with many of the usual trappings such as tax rates, random audits, penalties and occasional tax amnesties, as well as taxpayer heterogeneity. As we shall see, the introduction of risk aversion into the model and the resulting nonlinearity of the firm's utility function combines with the firm's dynamics and leads to a significant increase in complexity. This puts the problem finding the firm's optimal behavior well beyond the reach of analytical methods and requires powerful approximation techniques to be brought to bear.

The main contributions of this work are i) the use of the deep reinforcement learning techniques to obtain computational solutions for the firm's optimal behavior based on the Markov dynamical model of \citet{goumagias2012decision} and 
ii) a computational framework for exploring the behavior expected of self-interested risk-averse firms who may choose to engage in tax evasion in order to maximize their own utility. In addition, and on more practical grounds, we estimate the risk aversion coefficient of the ``average'' firm - or group of firms - given empirical data on its 
tax compliance and evaluate sample tax policies in terms of their benefit for the State (or, equivalently, the level of tax evasion they result in). 
To our knowledge, ours is the first work to apply deep learning in the context of taxation and tax evasion, and the first to obtain solutions that reveal the behavior of a risk-averse firm at a ``fine'' timescale, i.e., on a year-to-year basis, based on its evolving status in the ``eyes'' of tax authorities. 
We view our approach as particularly relevant both in light of the growing  interest  in deep learning applications and for the opportunities that our model affords to regulators in the design of effective policies that make entities behave more honestly.
The remainder of this paper is structured as follows. In Section~\ref{sec:litreview} we review the relevant literature and discuss how our approach is situated relative to previous work. Section~\ref{sec:model_desc} begins with a brief description of the tax system in which the firm operates, and explains the main parameters. In the same Section we describe a Markov-based model of the firm's evolution through the tax system and pose the main optimization problem we are interested in solving and the computational challenges involved. Our solution approach, combining Q-learning and Deep Neural Networks, is detailed in  Section~\ref{sec:approximate}.   Finally, Section~\ref{sec:evaluating} discusses the results we obtained -  using the Greek tax system as a case study for the sake of concreteness - and their relevance to the questions posed above regarding the firm's expected behavior, incentives for reporting profits, degree of risk aversion, and policy implications.

\section{Related work}
\label{sec:litreview}

Prior work related to optimal taxation and tax-evasion modeling can be grouped into two main categories: i) analytic (macroeconomic, and principle-agent based), and ii) computational (agent-based, simulation-based). The seminal work in the first category was~\citet{allingham1972income} who introduced a model of optimal taxation posed as a portfolio allocation problem.  Several scholars built on that model by also introducing labor supply ~\citep{yitzhaki1974income,baldry1979tax} and public goods offered ~\citep{cowell1981taxation}. The complexity of the phenomenon was highlighted early on by \citet{clotfelter1983tax} and \citet{crane1986inflation}, who challenged the monotonic relationship between tax rates and tax evasion. 
One of the drawbacks of the analytical approaches was that 
they often implied less behavioral heterogeneity on behalf of taxpayers than what was suggested by empirical evidence \citep{andreoni1998tax}, and - in order to remain tractable - they could not fully capture the dynamics of tax evasion  \citep{martinez2005multiple}.

In particular, beyond the issue of accounting for heterogeneity (e.g., in taxpayers'  risk-aversion), there exists much interesting structure in taxpayers' behavior if one considers ``fine-grained'' models of their evolution through the tax system. In that setting, one must reckon with the various random transitions the taxpayer may undergo from year to year, such as being audited, or offered the chance to participate in a tax amnesty program (we will provide details on such options shortly), or changing preferences via interaction with others. Such considerations have led to a number of recent computational-based approaches in the form of automaton-based \citep{garrido12} and agent-based \citep{gao09} models. Computational approaches may allow for more realism, by having, for example, a large number of agents interact with each other based on predetermined characteristics related to the taxation parameters and intrinsic utility functions \citep{pickhardt2014income}.  Their advantage is that they can offer empirically grounded and theoretically-informed policy implications, but they often suffer from a limited analytical tractability of the solutions they suggest. 

An attempt to overcome these limitations while modeling the year-to-year behavior of the firm was made by \citet{goumagias2012decision}, who introduced a parametric Markov-based model describing the evolution of a rational firm within the Greek tax system. The firm's goal was to maximize a discounted sum of its yearly after-tax revenues, possibly by engaging in tax evasion. That work showed that the firm would attempt to evade taxation as much as possible under the system currently in place, and produced ``maps'' showing which combinations of tax parameters lead firms to behave honestly and which do not. A severe limitation of \citet{goumagias2012decision} was the fact that it applied only to the special case of {\em risk neutral} entities. That assumption kept the firm's state and decision spaces conveniently small (it implied, for example, that the firm's optimal decision is to either be completely honest or to conceal as much profit as possible, eliminating ``intermediate'' options), making the problem of optimizing taxpayer behavior solvable via Dynamic Programming (DP) \citep{bertsekas1995dynamic}. Of course, most taxpayer entities are not likely to be risk neutral; thus, it becomes necessary to incorporate risk-aversion into the analysis in order to be able to predict the behavior of a broad spectrum of taxpayers and explore the effectiveness of tax policies in a more realistic setting. 

As we will discuss in Sec.~\ref{sec:challenges}, risk-aversion introduces nonlinearity in the firm's objective function,  making analytical or DP methods ineffective, and we will require some way of circumventing the curse of dimensionality in that context. Among the various alternatives, iterative dynamic programming can potentially allow for  tractable solutions 
\citep{jaakkola1994convergence}, however, that method's applicability is limited when faced with multiple sources of uncertainty, as is the case here. Computational solutions, including  artificial intelligence methods and neural networks for cost-to-go function approximation 
\citep{tsitsiklis1996feature,wheeler1986decentralized,watkins1989learning} will prove to be more promising in our setting.
Reinforcement learning-based methods, in particular, approximate the cost-to-go function via simulation and perform function approximation 
via  regression or neural networks \citep{gosavi2004reinforcement}. This 
approach includes algorithms  such as R-learning  
\citep{singh1994reinforcement,tadepalli1996scaling},  and Q-Learning 
\citep{sutton1998reinforcement,tsitsiklis1994asynchronous}.  One 
advantage of reinforcement learning which will be useful to us is that, unlike DP, 
the process can be set to update the value of the cost-to-go function for the states that are most often visited \citep{tsitsiklis1994asynchronous}.

Recently-proposed {\itshape deep learning } algorithms have greatly broadened the scope of applicability of artificial intelligence and machine learning, beyond ``classical" problems of pattern recognition \citep{lecun2015deep} and have shown great promise in approximating complex nonlinear cost-to-go functions \citep{schmidhuber2015deep}. To date, deep learning has been applied to challenging problems in areas including image recognition and processing \citep{krizhevsky2012imagenet}, speech recognition \citep{mikolov2011strategies}, biology \citep{leung2014deep}, analysis of financial trading \citep{krauss2017deep}, social networks \citep{perozzi2014deepwalk} and human behavior \citep{ronao16}. Here, we will make use of recent developments in deep reinforcement learning in order to obtain computational solutions for the firm's optimal behavior, with all of our model's complexities. 
This opens the door to more informed policy decisions by providing a computational platform for comparing tax policies (e.g., those with tax amnesty vs those without), estimating the firms' degree of risk aversion from empirical data, predicting the expected tax revenue for the government, or calculating the effects of a change in any tax parameter on revenues. %
\section{Model description}
\label{sec:model_desc}

We proceed with a brief discussion of the tax system within which the firm evolves, to be followed by the corresponding mathematical model. 
That model will be parametric, with many of the tax ``features'' commonly encountered, including random audits and penalties. 
Of course, when it comes time to make computations, we will have to select parameter values (tax rates, etc.) for a specific locale. 
We will focus on Greece in particular, for the sake of concreteness and because, with tax evasion being a significant and long-standing problem there, one can draw interesting and practical conclusions. 
However, the basic tax provisions we consider appear in most tax systems, and our model could be adjusted to describe matters in other countries as well.  

\subsection{A basic taxation system with occasional optional amnesties}
\label{sec:greece}
The basic components of our taxation system will
include - as is the case in most countries - a tax rate on profits, random audits for identifying tax evaders, and monetary 
penalties for under-reporting income. Those penalties, added to the original tax due on any unreported income discovered during an audit, will be proportional to the amount of unreported income and the time elapsed since the offense took place. We will also allow any penalty to be discounted somewhat for prompt payment. The tax authority will audit a small fraction of cases each year but will retain the right to audit a firm's tax returns for a number of years in the past. Any tax-evasion activity beyond that horizon will be considered to be beyond the statute of limitations.  

Our model will also include an optional tax amnesty in which the government may occasionally allow taxpayer entities to pay a fee in exchange for which past tax declarations are closed to any audits. This ``closure'' fee will be paid separately for each tax year a firm would like to exempt from a possible audit. 
It is worth noting that the appeal of tax amnesties as revenue collecting mechanisms is typically reinforced during and after long recessions \citep{Ross13,bayer14}. 
Amnesties are more commonly used than one might expect. For instance, only in the US, between 1982 and 2011, there were 104 cases of some form of tax amnesty \citep{Ross13}.  Other examples include India \citep{das1995tax}, and Russia \citep{alm1998tax}. In Greece, the closure option mentioned above was being offered roughly every 4-5 years during 1998-2006 (e.g., \citet{Mandate04} and \citet{Mandate08}). More recently, it was re-introduced in the Greek parliament with a new round being under consideration \citep{Mandate15}. The irregular usage of tax amnesties as tax revenue collection mechanisms increases the complexity of decision making both on behalf of the government and the taxpayer. 
The use of tax amnesties by firms essentially shrinks the audit pool. Thus, if in some year the government offers the closure option but a firm refuses to use it, that firm is more likely to be audited. For a more detailed explanation of the mechanics of closure, see \citet{goumagias2012decision}.
In practical terms, one question we would like to answer is whether such a measure (although it provides some immediate tax revenue) actually hurts long-term revenues because it might act as a counter-incentive to paying the proper tax \citep{bayer14}. 

\subsection{The behavior of risk-averse firms with optional closure}
\label{sec:model_eq}
The work in \cite{goumagias2012decision} codified the firm's time evolution through the tax system described above, in a compact Markov-based model which includes all of the basic features described in Sec.~\ref{sec:greece}, including tax rates, penalties, a five-year statute of limitations for audits of past tax statements, and occasional tax amnesty (closure). 
We will revisit it here briefly, in as compact form as possible, and extend it for our purposes. 

For a tax system with a five-year statute of limitations on auditing past tax statements, the firm's evolution can be described by the linear state equation \citep{goumagias2012decision}
\begin{equation}
    x_{k + 1} = Ax_{k} + Bu_{k} + n_{k},\
    \label{eq:firm_dynamics}
\end{equation}
where $x(0)$ is given, and $A \in \real^{7\times 7}$,  $B \in \real^{7\times 2}$, $n_k \in \real^7$ are as in \ref{ap:statedynamics}.

The firm's state at discrete time $k$ is given by the triple $x_k=[s_k, c_k, h_k^T]^T \in {\cal S} \times \{1,2\}, \times [0,1]^5$. Here, ${\cal S}$ is a 15-element set (in the discussion that follows, it will be convenient to use ${\cal S}=\{1,...,15\}$), containing the firm's possible tax statuses (see \citet{goumagias2012decision} for a graphical explanation): 
the first five elements of ${\cal S}$ correspond to the firm currently being audited, with 1-5 years since its last audit (any tax declarations ``older'' than 5 years are beyond the statute of limitations); elements 6-10 correspond to the firm using the closure option with 1-5 years having passed since its last audit or closure; and states 11-15 correspond to the firm being unaudited for 1-5  years (not being currently audited, nor using closure). Of the remaining state elements, $c_k$ is a two-level variable denoting whether the government has made the closure option available at time $k$, and $h_k$ contains the time history of the firm's past 5 decisions with respect to tax evasion, with elements in $h$ ranging from 0 (full disclosure) to 1 (the firm hides as much of its income as possible). 

In Eq.~\ref{eq:firm_dynamics}, $u_k$ is a 2-element vector containing the firm's actions in year $k$; the first element, $[u_k]_1 \in [0,1]$ denotes the fraction of profits that the firm decides to conceal, while the second, $[u_k]_2 \in \{0,1\}$ is a binary decision on whether or not to use the closure option, if it is available. 
In the term $n_k=[\omega_k, \epsilon_k, 0_{5 \times 1}]^T$, 
$\omega_{k}$ determines the first element of the ``next'' state vector, i.e., the firm's status in the tax system (e.g., being audited or not, or removing itself from this year's audit pool by making use of the closure option), according to a Markov decision process whose transition probabilities depend on the current state and the firm's decision to use closure (see \cite{goumagias2012decision}, also given in \ref{sec:appendixB} to facilitate review). The $\epsilon_{k}$ are Bernoulli-like, taking on the value 2 when the government offers the closure option (this is assumed to occur with some probability $p_0$), or 1 otherwise. 

The firm ``weighs'' its rewards (profit, plus any taxes it is able to save by declaring less of it) according to a constant relative risk aversion utility function
\begin{equation}
    U\left( z \right) = \frac{z^{1 - \lambda}}{1 - \lambda},    
    \label{eq:u}
\end{equation}
with $\lambda$ being the associated risk-aversion coefficient, and $z=g(x_k,u_k)$ being the reward the firm receives when in state $x_k$ and taking an action $u_k$. Based on the earlier description of the rules of the tax system, $g(\cdot,\cdot)$ is given by
\begin{multline}
    g\left( x_{k},\ u_{k} \right) = g\left( \lbrack s_{k},\ c_{k},\ h_{k}^{T} \rbrack^{T},\ u_{k} \right) \\ =
    R \cdot \begin{cases}
    \left( 1 - r + r\left\lbrack u_{k} \right\rbrack_{1} \right), & s_{k} \in \{ 11,\ \ldots,\ 15\} \\
     \left( 1 - r + r \lbrack u_{k} \rbrack_{1} - \ell \left( s_{k} - 5 \right) \right), & s_{k} \in \{ 6,\ \ldots,\ 10\} \\
     \begin{pmatrix}
    1 - r + r\left\lbrack u_{k} \right\rbrack_{1} - r\sum_{i = 1}^{s_{k}}\left\lbrack h_{k} \right\rbrack_{6 - i} \\ -
    r \beta_{d}\beta \sum_{i = 1}^{s_{k}}{i\left\lbrack h_{k} \right\rbrack}_{6 - i} \\
    \end{pmatrix}, & s_{k} \in \{ 1,\ \ldots,\ 5\} \\
    \end{cases}
    \label{eq:policy_uk}
\end{multline}
where $R$ denotes the firm's annual revenues, $r$ is the tax-rate, $\ell$ the closure cost (paid if the firm decides to take advantage of that option in the event it is offered), $\beta$ the tax-penalty and $\beta_d$ is the discount factor for prompt payment.
In Eq.~\ref{eq:policy_uk}, the top term corresponds to the firm's reward if it is not audited, so that depending on $[u_k]_1$, it may pay all to none of the tax due. In the middle term, the firm is using the closure option, so that it pays $\ell$ for as many years as it has gone unaudited, up to a maximum of five. Finally, the bottom term in Eq.~\ref{eq:policy_uk} corresponds to the firm being audited, so that it pays any back taxes due (based on its historical behavior) and the corresponding penalties, as per our earlier description.

The firm is assumed to act in a self-interested way and thus chooses its policy $u_k$ so as to maximize the discounted expected reward:
\begin{equation}
    \max_{u_k} \mathcal{E}_{\omega_{k},\ \epsilon_{k}}\left\{ \sum_{k = 0}^{\infty}{\gamma^{k}U\left( g(x_{k},u_{k}) \right)} \right\}    
    \label{eq:discexprew}
\end{equation}
where $\gamma \in (0,1)$ denotes the discount factor.

It can be shown (in a way similar to \citet{goumagias2012decision}) that the Bellman equation whose solution maximizes (\ref{eq:discexprew}) is equivalent to 
\begin{multline}
    J_{\infty}\left( i,q, h \right) = \max_u \{ U\left( g\left( i,q,h,u \right) \right) \\ + \gamma\sum_{t = 1}^{2}{\sum_{j = 1}^{15}{P_{\text{qji}}\left( \left\lbrack u \right\rbrack_{2} \right)Pr(\epsilon = t)J_{\infty}(j,t,Hh + [0~0~0~0~1]\left\lbrack u \right\rbrack_{1})}} \}    
    \label{eq:bellman_eq2}
\end{multline}
where, for
convenience, we have slightly abused the notation by writing $J_\infty(i,q,h)$ instead of $J_\infty(x)$, with
$i \in S = \{1, \ldots,15\}$, $q \in \{1,2\}$, and $h \in \lbrack 0,1 \rbrack^{5}$.

\subsection{Challenges in solving for the firm's expected strategy} %
\label{sec:challenges}
There is a significant difficulty when it comes to solving Eq.~\ref{eq:bellman_eq2} for the optimal firm reward (and the associated tax-evasion policy), stemming from the continuity of certain elements in the state and control vectors. As we have already mentioned, the first element, $\lbrack u_{k} \rbrack_{1} \in \lbrack 0,1\rbrack$, of the control vector $u_{k}$ denotes tax-evasion as fraction of the firm's annual revenues. This implies that $u$ as well as $x$ are continuous because the firm's last five tax-evasion decisions are always incorporated into the state. This makes Eq.~\ref{eq:bellman_eq2} difficult to compute.

One may attempt to circumvent this problem by discretizing the variables in question to render both the state and the control vector discrete. For example, we may instead consider $\lbrack u_{k} \rbrack_{1} \in \lbrack 0, 0.01, \ldots, 0.99, 1 \rbrack$, and assume that tax-evasion takes place in increments of 1\%, which seems like a reasonable level of coarseness. However, after thus discretizing the control and state spaces, the number of state-control pairs, ($x,u$), remains large. Specifically, we are left with $15 \times 2 \times 101^{5} \times 202$ potential pairs (the number of the elements of the state vector $x_k$ including all possible combinations of control for the past five years, times the number of possible controls in $u_k$). Such a number of states is too large for DP to be effective in solving the stationary Bellman equation via value iteration, for example, 
because: i) ``visiting'' every state in order to update the value function associated with Eq.~\ref{eq:bellman_eq2} becomes infeasible and ii) it is difficult to even store the function $J(x,u)$ (the value of applying decision $u$ while at state $x$, as a precursor to computing the maximum in the above equation) in tabular form, as one would have to do if Eq.~\ref{eq:bellman_eq2} were to be solved via value iteration, for example.

The work in \citet{goumagias2012decision} circumvented these difficulties by assuming risk-neutrality ($\lambda=0$) on behalf of the firm (and thus linearity of the reward function) and successfully applied DP after determining that $\lbrack u_{k} \rbrack_{1}$ should only take a ``bang-bang'' form (conceal as much revenue as possible or none at all), leading to a significant reduction in the number of state-control pairs. In our case, however, the cost\emph{-to-go} function (Eq.~\ref{eq:policy_uk}) is non-linear, so that we must consider the full range of control values, and it is thus computationally difficult to apply DP.

One way to go forward is to combine: i) an approximation method to estimate the value function $J_{k}$ and ii) an approximate way of storing the optimal values of $J_{k}$, based on the optimal policy. To address the former we will use reinforcement learning -- specifically Q-learning, as described in~\citet{sutton1998reinforcement}, where $J_{k}$ will play the role of the Q-function $Q(x_k,u_k)$, while for the latter, a deep Artificial Neural Network will be used, as we will discuss shortly.
\section{Constructing an approximator: Deep Q-Learning}
\label{sec:approximate}
We experimented with various choices of learning algorithms and neural network architectures for the purposes of learning and storing the optimal value function given in the previous Section. In the following we describe our solution, combining Q-learning and a Deep Neural Network, and discuss some of the difficulties involved and how they can be overcome.

\subsection{Q-learning}
\label{sec:Q}
Q-learning is a model-free reinforcement learning method \citep{sutton1998reinforcement}, that is used to find an optimal action-selection policy for any given finite MDP. 
In the ``language" of \citet{sutton1998reinforcement}, an agent (in our case the firm) observes the current state $x_k \in \mathcal{X}={\cal S} \times {\cal C} \times [0,1]^5$ at each discrete time step $k$, chooses an action $u_k \in \mathcal{U}=[0,1] \times \{0,1\}$ according to a possibly stochastic policy $\pi$, mapping states to actions, observes the reward signal $U\left( g\left(x_{k}, u_{k} \right) \right) \in \mathbb{R}$, and transitions to a new state $x_{k+1}$.
The objective is to maximize an expectation over the discounted return, as in Eq.~\ref{eq:discexprew}.

Briefly, Q-learning involves sequentially updating an approximation of the action-value function, i.e., the function that produces the expected utility of taking a given action at a given state and following the optimal policy thereafter.
The so-called $Q$-function of a policy $\pi$ is $Q^{\pi}(x,u) = {\cal E} \left\{  D_k | x_k = x,u_k = u \right\}$,
where 
\begin{equation}    D_k =  \sum_{i=0}^\infty \gamma^{i} U\left( g\left(x_{k+i}, u_{k+i} \right) \right),
\end{equation}
and the state evolution proceeds under the policy $\pi$.
Finally, the optimal action-value function $Q^{*}(x,u) = \max_{\pi} Q^{\pi}(x,u)$ to which the learning process is to converge, obeys the Bellman Eq.~\ref{eq:discexprew}.

For our purposes, in the notation of Sec.~\ref{sec:model_desc}, the function $J$ we are seeking (\ref{eq:bellman_eq2}) is simply the $Q^*$ function, after having maximized over $u$.
Common choices for modeling the Q-function are lookup tables and linear approximators, among others. However, these models suffer from poor performance and scalability problems, and cannot possibly handle the high-dimensional state space involved in our case, as we discussed in Sec.~\ref{sec:challenges}. An efficient alternative to the aforementioned models are neural networks.

\subsection{Deep Q-Networks (DQN)}

Deep $Q$-learning (DQN) was introduced by \citet{mnih2015human}, and uses neural networks parametrized by $\theta$ to represent $Q(x,u;\theta)$, where the $Q$ function is augmented with a parameter vector $\theta$, usually consisting of the weights and biases of the multiple layers of the network.
Neural networks, viewed as general function approximators, are trained ``end-to-end", and can efficiently handle high-dimensionality problems. 
Recently, a DQN surpassed human performance in 49 different Atari games~\citep{mnih2015human}.
For our purposes, the DQN will receive as input the firm's state $x_k$ and will have to produce the optimal decision, $u_k$. Because the network will be trained to capture the optimal firm policy, we will sometimes refer to it as the ``policy network''.

DQNs are trained iteratively using stochastic gradient descend, until convergence. This is done by minimizing, at each iteration~$i$, a loss function of the network's parameters, $\mathcal{L}_i$, which is expressed as
\begin{gather}
\mathcal{L}_i(\theta_i) = {\cal E}_{x,u,r,x'} \left\{ \Delta Q^{2} \right\}, \mbox{with}\\
\Delta Q = Y^{DQN} - Q(x,u; \theta_{i}), \mbox{and}\\
Y^{DQN} = U\left( g\left(x_{k}, u_{k} \right) \right) + \gamma \max_{u'} Q(x',u';\theta_{i}^{-}),
\label{eq:loss}
\end{gather}
and $\theta_{i}^-$ is an ``older'' copy of the network's parameters, as we explain next.
Function approximation using neural networks can be unstable, and we observed such behavior in our numerical experiments, particularly after we introduce a second source of uncertainty in the form of closure availability. 
Following \citet{mnih2015human}, to stabilize the process we use a so-called ``target network", i.e., a copy of our original DQN which has the same architecture but a different set of parameters, $\theta_{i}^{-}$. The parameters of the target network represent an older version of the policy network and are updated at a slower rate. 
Thus, while the policy network acts to produce inputs $u$ that will steer the firm to its next state, the slowly-updated target network is used to compute $Y^{DQN}$ which, in turn, is used to improve the parameters of the policy network
via gradient descent:
\begin{equation}
\nabla_{\theta_i} \mathcal{L}_i(\theta_i) = {\cal E}_{x,u,r,x'} \left\{ \Delta Q \nabla_{\theta_i} Q(x,u; \theta_{i}) \right\}.
\label{eq:loss_grad}
\end{equation}

While training the DQN, we must choose an action $u$ to drive the state at each iteration. That action is to be chosen from $Q(x,u; \theta_{i})$ using an $\epsilon$-greedy policy that selects the $u$ that maximizes  $Q$ with probability  $1-\epsilon$, or a random $u$ with probability $\epsilon$.
Additionally, our DQN uses so-called ``experience replay''~\citep{lin1993reinforcement}. During learning, we maintain a set of episodic experiences (tuples that include the state, the action taken, the resulting state and reward received). The DQN is then trained by sampling mini-batches of those experiences. This has the effect of stabilizing the learning process and avoids overfitting. Experience replay was used very successfully by \citet{mnih2015human} and it is often motivated as a technique for reducing sample correlation, while also enabling re-use of past experiences for learning. Furthermore, it is a valuable tool for improving sample efficiency and can also improve performance by a significant margin, as it did in our case.

A final but important modification was the use of Double Q-learning, a technique introduced very recently by \citet{van2016deep}. Double Q-learning for DQN (DDQN)  reduces overestimation of the Q-values by decomposing the max operation in the target network into action selection and action evaluation. 
Thus, instead of using the target network's maximum Q-value estimate in Eq.~\ref{eq:loss}, we use the target network's Q-value of the current network's best action.
The DDQN update equations are the same as for DQN, after replacing the target $Y^{DQN}$ in Eq.~\ref{eq:loss} with
\begin{equation}
Y^{DDQN} = U\left( g\left(x_{k}, u_{k} \right) \right) + \gamma Q(x', \argmax_{u'}Q(x',u';\theta_{i});\theta_{i}^{-}).
\end{equation}
The entire Double DQN training loop is given in pseudocode in Algorithm~\ref{alg:dqn} below. 
\begin{algorithm}[h]
\SetAlgoNoLine
  $\triangleright$ Initialize experience replay memory $D$, action-value function $Q$ with random weights $\theta$ and set $\theta^{-} = \theta$.\\
      
  \For{episode = 1 to $M$}{
      \For{k = 1 to $K$}{
          \tcp{Take an action}
            Select $u_k$ randomly with probability $\epsilon$ else $\argmax_{u}Q(x_k, u; \theta)$\\
          Execute $u_k$ and observe reward $U\left( g\left(x_{k}, u_{k} \right) \right)$ and state $x_{t+1}$\\
          Store transition $(x_k, u_k, U\left( g\left(x_{k}, u_{k} \right) \right), x_{t+1})$ in $D$\\
          \tcp{Training step}
          Sample minibatch $(x_j, u_j, r_j, x_{j+1})$ from $D$\\
          $Y^{DDQN} = U\left( g\left(x_{j}, u_{j} \right) \right) + \gamma Q(x', \argmax_{u'}Q(x',u';\theta_{i});\theta_{i}^{-}) $\\
          Perform a gradient descent step $\nabla_\theta(Y^{DDQN} - Q(x_j, u_j; \theta))^2$\\
          \tcp{Update target network}
          Every $C$ steps reset target network, i.e., set $\theta^- = \theta$
      }
  }

\label{alg:dqn}
\caption{Double DQN \citep{van2016deep}}
\end{algorithm}

\subsection{DQN architecture}

Our network architecture was inspired by the model of \citet{mnih2015human}. The action-space described in Sec.~\ref{sec:model_desc} consists of two action elements $[u]_1$ and $[u]_2$.
The firm's tax-evasion level is determined by $[u]_1 \in [0,1]$, discretized in intervals of $1\%$ resulting a set of $101$ actions. This convention is commonly used to take advantage of the off-policy stability of Q-learning compared to on-policy $SARSA-\lambda$, actor-critic or policy gradient approaches.
The firm's use of the closure option is $[u]_2 \in \{0,1\}$, and if closure is not available then $[u]_2 = 0$.

\begin{figure}[tp]
    \centering
    \includegraphics[width=0.4\linewidth]{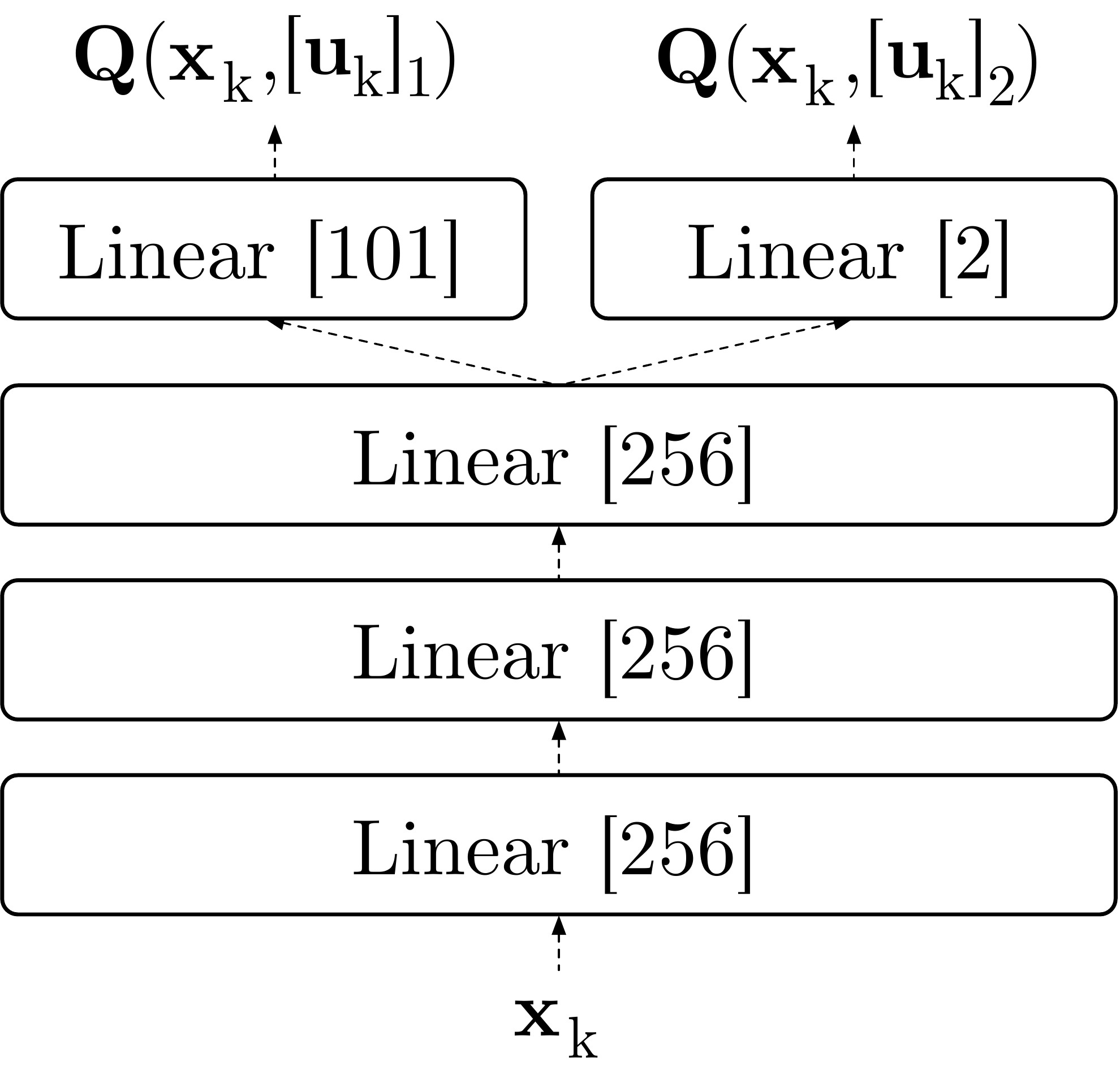}
    \caption{Schematic representation of our DQN. We use a 4-layer network that takes as input the current state $x_k$. The first three layers consist of $256$ neurons, followed by two parallel layers of $101$ and $2$ neurons, for computing $Q(x_k, [u_k]_1)$ and $Q(x_k, [u_k]_2)$, respectively.}
    \label{fig:architecture}
\end{figure}

Our approximator (see Fig.~\ref{fig:architecture}) is a 4-layer multilayer perceptron (MLP) and takes as input the current state $x_k$.
The first three layers consist of $256$ neurons, followed by two parallel linear layers of $101$ and $2$ neurons, for computing $Q(x_k, [u_k]_1)$ and $Q(x_k, [u_k]_2)$, respectively. The network makes use of the rectified linear unit (ReLU) transformation function $f(x) = \max(0,x)$ between  layers. 

Finally, our setting requires the DQN to produce two action elements $([u]_1, [u]_2)$. To improve the scalability of our approximator, and after numerical experimentation, we opted to use independent Q-learning to learn two different Q-functions (one for each component of the firm's decision, $[u]_1$ and $[u]_2$, as 
in \citep{narasimhan2015language,foerster2016learning}. In this case, the DQN loss is expressed as
\begin{gather}
\Delta Q_{[u]_1} = U\left( g\left(x_{k}, u_{k} \right) \right) + \gamma max_{[u_{k+1}]_1} Q(x_{k+1}, [u_{k+1}]_1) - Q(x_k, [u_k]_1)\\
\Delta Q_{[u]_2} = U\left( g\left(x_{k}, u_{k} \right) \right) + \gamma max_{[u_{k+1}]_2} Q(x_{k+1}, [u_{k+1}]_2) - Q(x_k, [u_k]_2)\\
\mathcal{L}_i(\theta_i) = {\cal E}_{x,u,r,x'} \left\{ \Delta Q_{[u]_1}^{2} + \Delta Q_{[u]_2}^{2} \right\}.
\label{eq:loss_features}
\end{gather}

\section{Evaluating the model: results and discussion}
\label{sec:evaluating}

As we have mentioned in the Introduction, we are generally interested in being able to evaluate the firm's decisions (assuming that it acts in a self-interested way) - and maximum expected utility under various degrees of risk-aversion, thereby producing a tool that could be used to predict firm behavior, compute tax revenue, and to gauge the reaction of the firm to tax policy scenarios under consideration by the government. We are also interested in characterizing the firm's strategy by determining, for example, whether the firm would be expected to use a constant degree of tax-evasion ($[u]_1$) in every state (as in \citet{goumagias2012decision}), finding the firm's coefficient of risk-aversion given empirical estimates of the degree of tax evasion, and examining whether it is beneficial for the government to offer the closure option in any of the settings discussed in the Introduction. 

\subsection{Model parameters and Training setup} %
\label{sec:trainsetup} 

The various tax parameters present in our model were selected using Greece as a case study for the sake of concreteness, to facilitate comparisons with prior work \citep{goumagias2012decision}, and because that country presents an interesting case as it is plagued by widespread tax evasion (we will discuss estimates in Sec.~\ref{sec:estimates_risk}). 
Specifically, the tax and audit rates were $r=0.24$ and  $p=0.05$, respectively; the statute of limitations for auditing past tax statements was 5 years; the penalty for underreported profit was $\beta=0.24$ (24\% annually); potential tax penalties were discounted by 40\% if paid immediately ($\beta_d=0.6$); and, finally, the cost for the firm to use the closure option - if available - was $\ell=0.023$. 

Training our DQN-based model to optimize the firm's behavior for any one set of parameters (risk-aversion coefficient, closure probability and cost, audit probability, penalty coefficient) required about 2 days on  an Intel\textsuperscript{\textregistered} Xeon\textsuperscript{\textregistered} X5690 CPU with 72GB of RAM. Our source code is freely available under an open-source license at \url{https://github.com/iassael/tax-evasion-dqn}. 
The network was trained on $50,000$ episodes of the firm's evolution, each lasting $250$ time steps. The network's performance was evaluated every $100$ episodes as the average discounted reward of those episodes.
We followed the training methodology proposed by \citet{mnih2015human}, using  Double Q-Learning \citep{van2016deep}.
Because $x_k \in [0,1]^{21}$, the inputs to the network were ``shifted'' by subtracting $0.5$ from all elements of the state $x_k$. Shifting the inputs to be evenly spread around $0$ resulted in faster convergence\footnote{A simple example where this type of shifting improves learning is the case of one-hot encoded inputs $x$, where both the weights $W$, and biases $b$, of the network can be being ``learned'' even when the original inputs are zero, i.e., $f(x) = ReLU(Wx + b)$, whereas without shifting, only $b$ would be learned when $x=0$.}.

As usual, the network's training objective was to minimize the mean squared temporal difference error. Thus, the backpropagated gradients described above were significantly affected by the scale of the rewards. Looking at the form of the risk-averse utility function $U( \cdot )$ in Eq.~\ref{eq:u}, this becomes problematic for input values close to $0$, where $U$ dives to $-\infty$. 
To stabilize the training process numerically, the values returned by $U$ were clipped below, so that they always lie in $[-1, 0)$. That is, if the argument of $U$ was less than $\epsilon_{thresh}$, where $U(\epsilon_{thresh})=-1$, the argument was replaced by $\epsilon_{thresh}$.
Our empirical evaluation showed that reward clipping was crucial to deal with the steep non-linear scale of rewards. The particular value of -1 was not critical - more negative values work just as well, as long as they are ``far'' from the utility values the firm usually operates around, but not too negative so as to end up in extremely steep parts of $U$ near zero. 

Our $\epsilon$-greedy exploration policy used $\epsilon=0.5$ which linearly decreased to $\epsilon=0.1$ in the first $5000$ episodes. This resulted in a highly-explorative policy in the beginning which rapidly converged to a more exploitative one. 
The training process took advantage of past experiences, as we explained above (experience replay with mini-batches of size $100$), and the target network described in Sec.~\ref{sec:Q} was updated every $10$ episodes. The networks' parameters were optimized using  Adam~\citep{kingma2014adam} with a learning rate of $10^{-4}$.

We proceed by first evaluating our model in the case of a risk neutral firm - for the purposes of comparison with prior work. Following that, we will discuss the case of a risk-averse firm and will explore its behavior.

\subsection{Risk-neutral firms: comparison with known optimum.}
\label{sec:validation}
Before attempting to compute a risk-averse firm's expected behavior, we validated our approach against the known optimal solution for risk-neutral firms from \citet{goumagias2012decision}. Tab. \ref{tab:validation} shows the firm's total discounted rewards in four cases which are of interest, according to how often the closure option is offered each year: a) never, b) with probability 0.2, c) always, and d) periodically, every 5 years.

\begin{table}[htpb]
\centering 
\begin{tabular}{|l|l|l|}
\hline
Closure Option & Dynamic Programing & DQN \\
\hline
Never & 3254.6    & 3270.66 \\
\hline
$p_{closure}=0.2$ & 3307.9    & 3316.76 \\
\hline
Always & 3358.3     & 3357.01 \\
\hline
5-periodic & 3319.7    & 3335.75 \\
\hline
\end{tabular}
\caption{Total discounted revenue for risk-neutral firm, as computed by our model vs. via Dynamic Programming as reported in \citet{goumagias2012decision}.}
\label{tab:validation}
\end{table}
Our DQN approach is inherently an approximate one. We note however that the firm revenues we computed differ less than 0.5\% from the ``true'' values computed via DP. Besides the optimal firm revenues, the optimal firm policies were identical to those found in \citet{goumagias2012decision} in each of the four cases examined, i.e., it was always optimal for the firm to conceal as much profit as possible and to make use of the option whenever available. 

\subsection{The behavior of risk-averse firms - ranking sample tax policies}
\label{sec:risk-averse-results}

We performed a series of runs designed to explore the effect of risk aversion on the behavior of the firm, by keeping the tax-parameters fixed to the values mentioned in Sec.~\ref{sec:trainsetup}, and varying the firm's risk aversion coefficient, $\lambda$ from 0 to 7 in steps of 1, for each of the four scenarios of interest with respect to the availability of closure (never, 20\% of the time, always, every 5 years). 

The first notable difference with the risk-neutral case \citep{goumagias2012decision} is that the optimal degree of tax-evasion, $[u]_1$, for $\lambda>0$ was {\em not} constant.
That is, in every case, our DQN-based approach
converged to a state-dependent (static) policy which achieved a higher average utility than would have been possible using any constant value for $[u]_1 \in [0,1]$ (meaning that the same value of $[u]_1$ would be used regardless of which state we were in). See Tab.~\ref{tab:compare_constant} for a comparison in the case where $\lambda=2.6$ (we have chosen this particular value because it will be of special interest in Sec.~\ref{sec:estimates_risk} - similar results hold for different values of $\lambda$).
\begin{table}[htpb]
\centering 
\begin{tabular}{|l|l|l|}
\hline
Closure Option & Max. discounted  & Max. discounted utility \\
 &  utility (average $[u]_1$) & with constant $[u]_1$ \\  
\hline
Never &  -1.91474$\cdot 10^{-2}$ ~~(0.29)
    & -1.98007$\cdot 10^{-2}$
 ~~(0.21) \\
\hline
$p_{closure}=0.2$ & -1.87780$\cdot 10^{-2}$ ~~(0.40)
    & -1.94671$\cdot 10^{-2}$
 ~~(0.31) \\
\hline
Always & -1.40147$\cdot 10^{-2}$ ~~(1)
 & -1.40147$\cdot 10^{-2}$
 ~~(1) \\
\hline
5-periodic & -1.86345$\cdot 10^{-2}$
 ~~(0.43)
 & -1.89893$\cdot 10^{-2}$
 ~~(0.37) \\
\hline
\end{tabular}
\caption{Long-term discounted expected utilities for a risk-averse ($\lambda=2.6$) firm: maximum achieved vs. maximum under the best {\em constant} $[u]_1$. The numbers in parentheses indicate the time-average value of $[u]_1$ leading to the maximum expected utility, and the optimum {\em constant} $[u]_1$, respectively.}
\label{tab:compare_constant}
\end{table}

In terms of the four tax policies under consideration, we observe from Tab. ~\ref{tab:compare_constant} that - as in the risk-neutral case - the firm obtains a higher maximum discounted utility when the closure option is offered more frequently or more predictably. This implies that, from the point of view of government, the tax revenue collected is highest when the closure option is never offered at all. We will have more to say about this in Sec.~\ref{sec:policy_implications}.

Regarding the use of closure by the firm ($[u]_2$) we found that, 
for the tax-parameters currently in use,
if the closure option is always offered then the firm must always take advantage of it (so that it is never audited). If the option is offered stochastically or every five years, then it is optimal for the firm to use it {\em unless} the firm has {\em just} been audited (this being a departure from the optimal risk-neutral policy). With respect to the level of tax-evasion, $[u]_1$, the fact that the optimal policy is not constant makes it difficult to characterize it in a ``compact'' way, especially when closure is offered stochastically or periodically. We will discuss ways of exploring the structure of $[u]_1$ later in this Section.

\subsection{The effect of risk aversion on tax evasion}
\label{sec:risk_effect}
\begin{figure}[htb]
    \centering
    \includegraphics[width=0.45\linewidth]{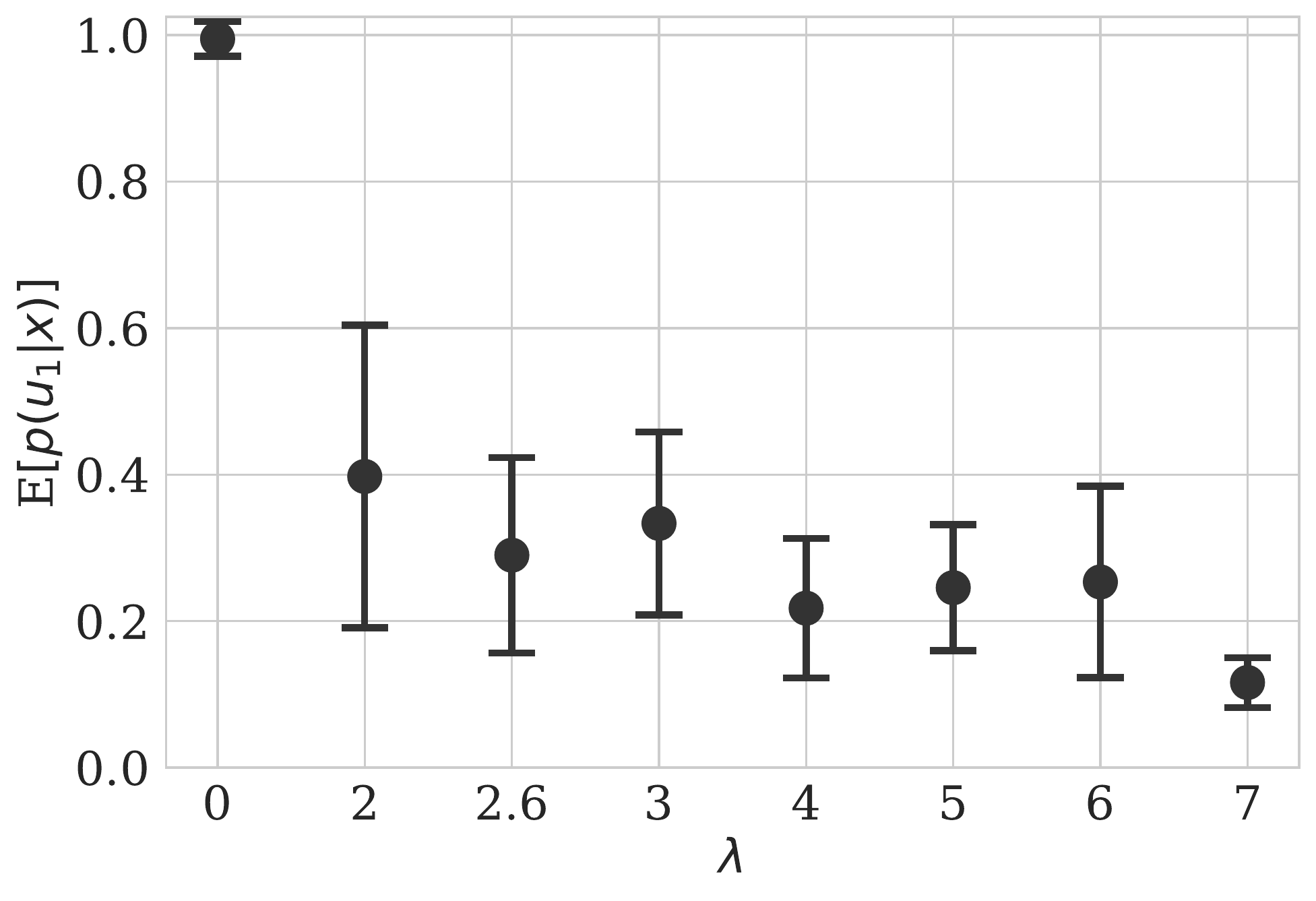}
    \includegraphics[width=0.45\linewidth]{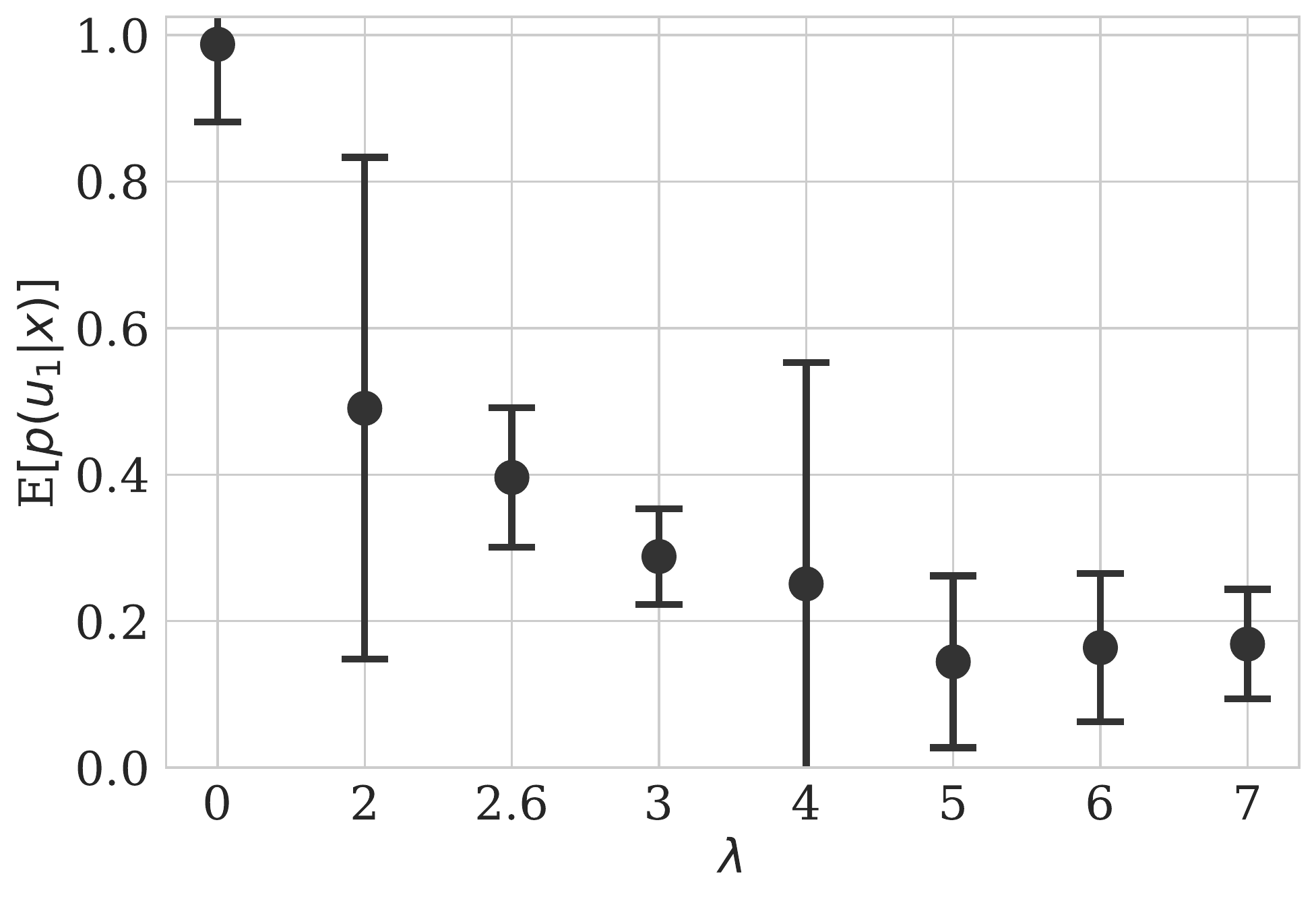}\\
    \qquad \quad$p_{closure}=0$ \quad\quad \qquad \qquad \qquad \qquad \quad $p_{closure}=0.2$ \\
    \vspace{1cm}
        \includegraphics[width=0.45\linewidth]{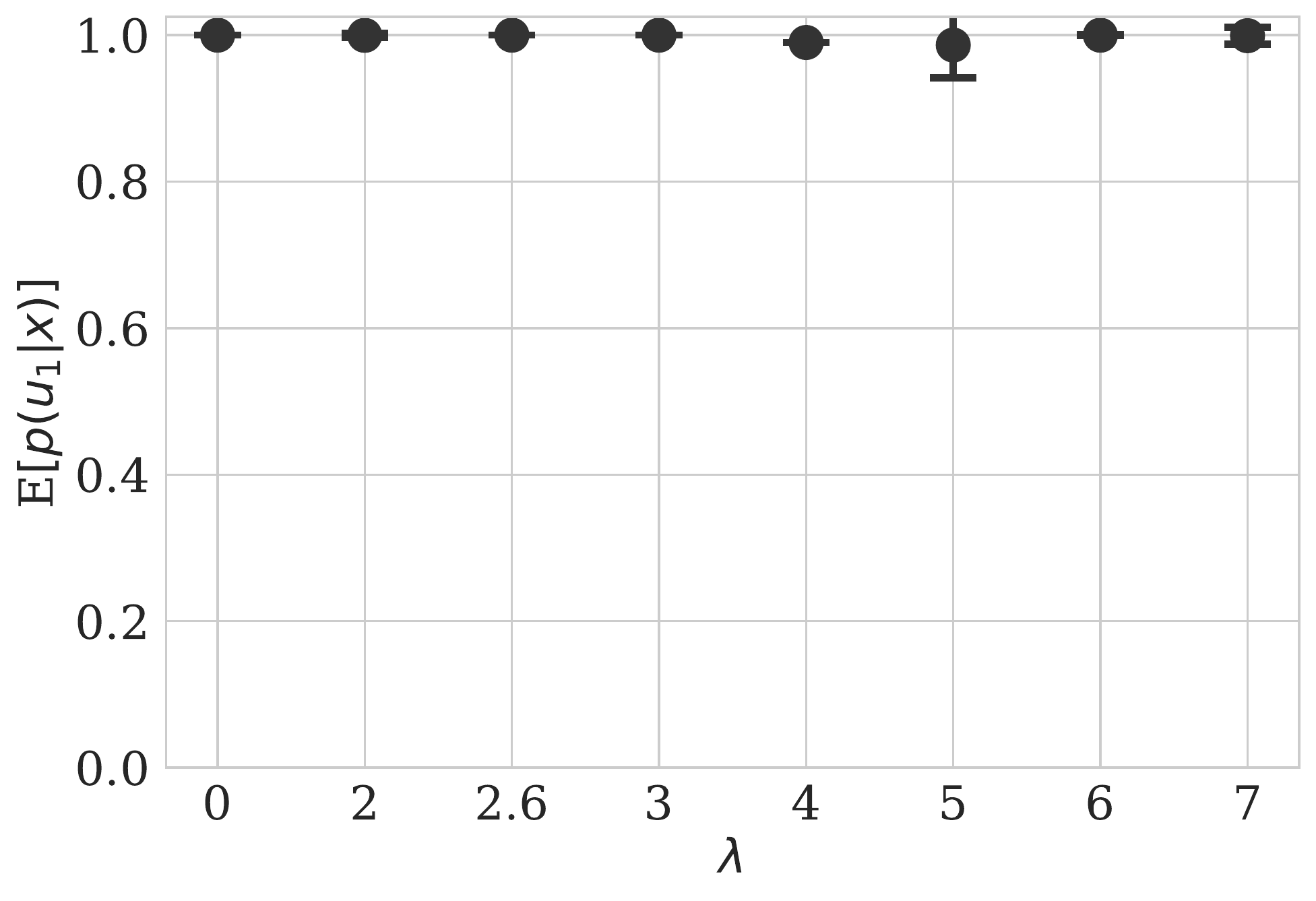}
    \includegraphics[width=0.45\linewidth]{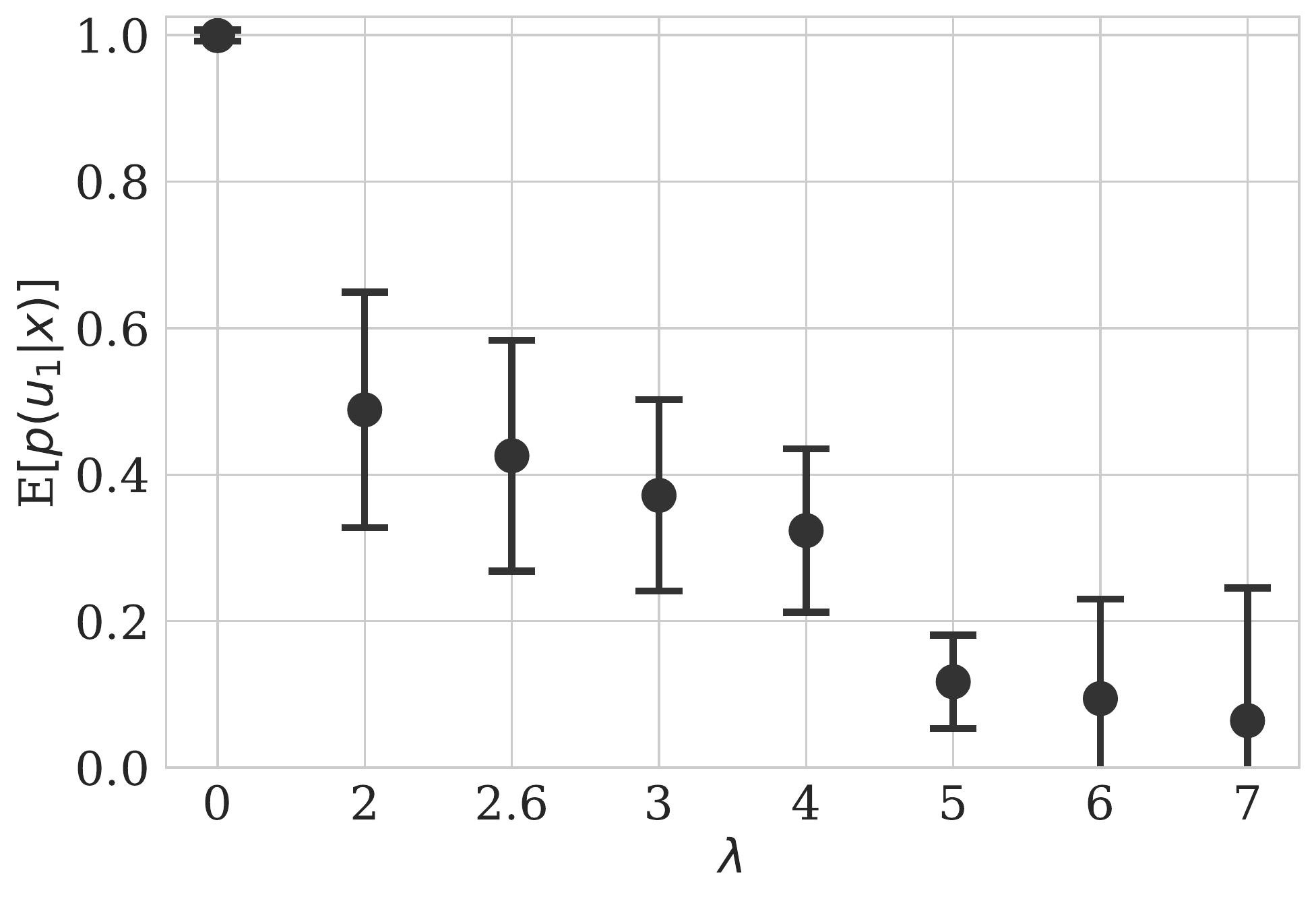}\\
    \qquad \quad Always\quad\quad \qquad \qquad \qquad \qquad \qquad\quad 5-periodic \\
    \caption{Average level of tax-evasion over the firm's evolution ($100$ episodes of $250$ time steps each). Dots represent  mean values, bars indicate $\pm$ one standard deviation.}
    \label{fig:k1}
\end{figure}

To gain insight into the firm's behavior we plotted the {\em average} $[u]_1$ over the course of the firm's lifetime against the firm's risk-aversion coefficient, $\lambda$. 
Fig.~\ref{fig:k1} shows the rate at which the average level of tax evasion ($[u]_1$) declines as the firm becomes more risk-averse, for each of the four scenarios regarding the availability of closure, where for each value of $\lambda$ there were $100$ episodes executed with $250$ time steps each.  
The approximate nature of our approach comes through in the fact that in the case where closure is never offered (Fig.~\ref{fig:k1} - top left), there are times where the average level of tax-evasion increases as $\lambda$ (the firm's risk aversion) increases, although we expect the opposite to occur. There is, however, a clear downward trend in the vast majority of cases showing that as the firm becomes more risk averse (higher $\lambda$) the firm becomes more ``honest'' on average. 
It is also worth mentioning that it is not trivial to obtain high numerical precision with an approximation method such as  ours when the utility function is highly nonlinear (i.e., in our case, very steep near zero where the firm would find itself if it had to pay a penalty at audit time, and relatively ``flat'' for values of income associated with non-audit states). One possible solution for learning value-functions over different reward ``scales'' is offered in \citet{van2016learning}; however, the implementation is complex, hence we opted for reward clipping as discussed in Sec.~\ref{sec:trainsetup}.

\subsubsection{Calculating the risk aversion of Greek firms}
\label{sec:estimates_risk}
In Fig.~\ref{fig:k1} we included data points for $\lambda=2.6$ on the horizontal axis. That value of the risk-aversion coefficient is significant because (see Fig.~\ref{fig:k1} top-right) it leads to a 40\% average tax-evasion on behalf of the firm.  It was identified by numerical experimentation, essentially using bisection on $\lambda$ to make the average $[u]_1=0.4$. 
As we have mentioned before, the 40\% level is reported in the literature as the estimated tax-evasion level in Greece \citep{artavanis2016measuring}, and so our approach allows us to estimate the risk aversion coefficient of the average Greek firm (or to re-estimate it for all or a subset of firms, as newer empirical data becomes available). 

\subsection{Exploring the optimal policy for a representative firm ($\lambda=2.6$)}
\label{sec:lam26}
As we have seen, the firm's optimal policy is not constant in three of the four closure availability scenarios (the exception is the case where closure is {\em always} available, where it is always best to conceal all profit). 
Because of the complexity of the problem and the large number of states ($15 \times 101^5$), it is difficult to represent or even visualize the optimal policy in a compact form. 
We have thus attempted to gain insight by examining the statistics of $[u]_1$ and $[u]_2$ and by using decision trees, 
as well as various projections of the state-to-decision mapping encoded in the DQN that are of practical relevance because they reveal how the tax evasion level is related to i) the tax status of the firm  (i.e., how many years since its last audit or closure), and ii) the amounts that the firm has previously concealed but are still within the statute of limitations in the event of an audit. 

\begin{figure}[htbp]
    \centering
    \includegraphics[width=0.45\linewidth]{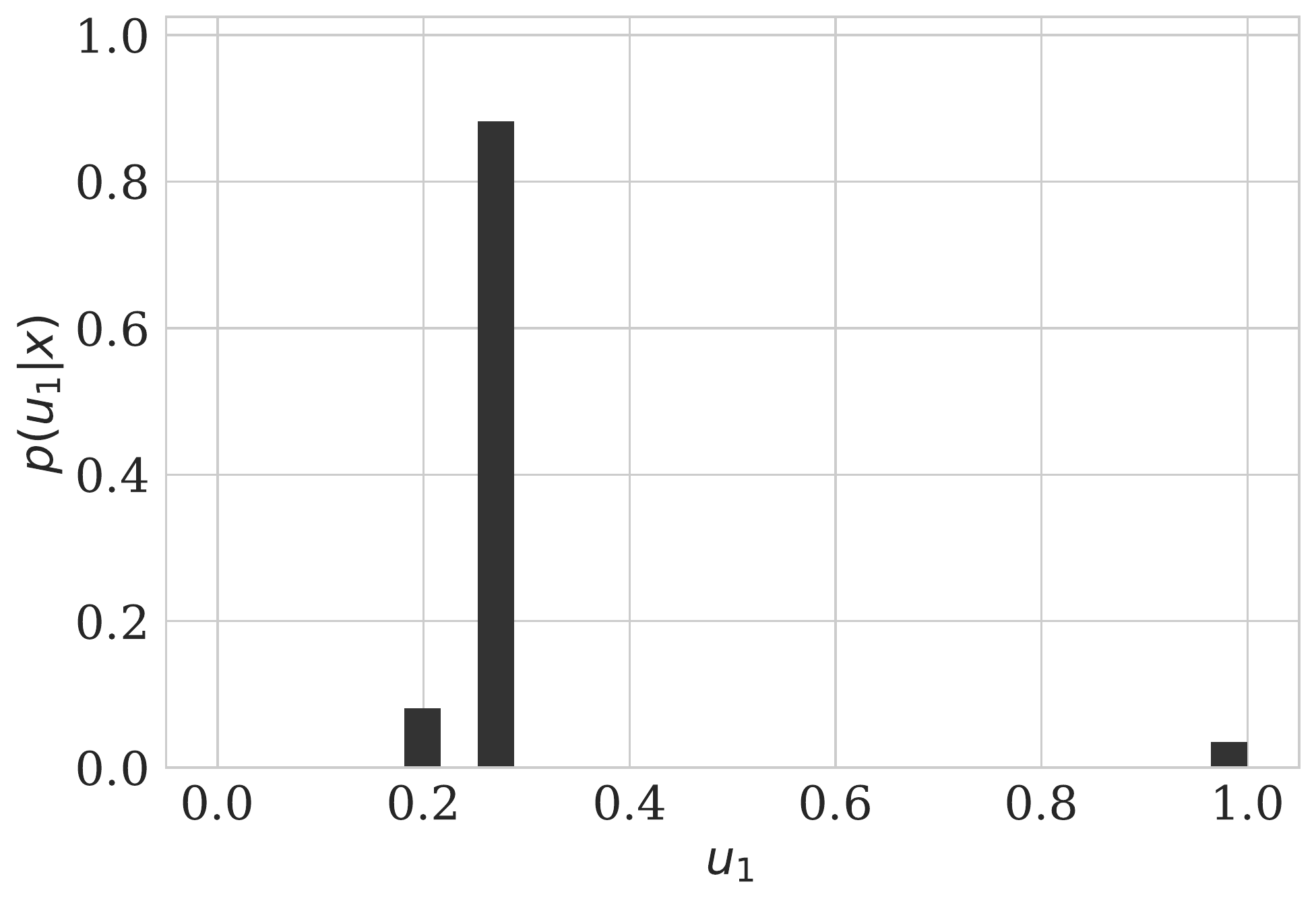}
    \includegraphics[width=0.45\linewidth]{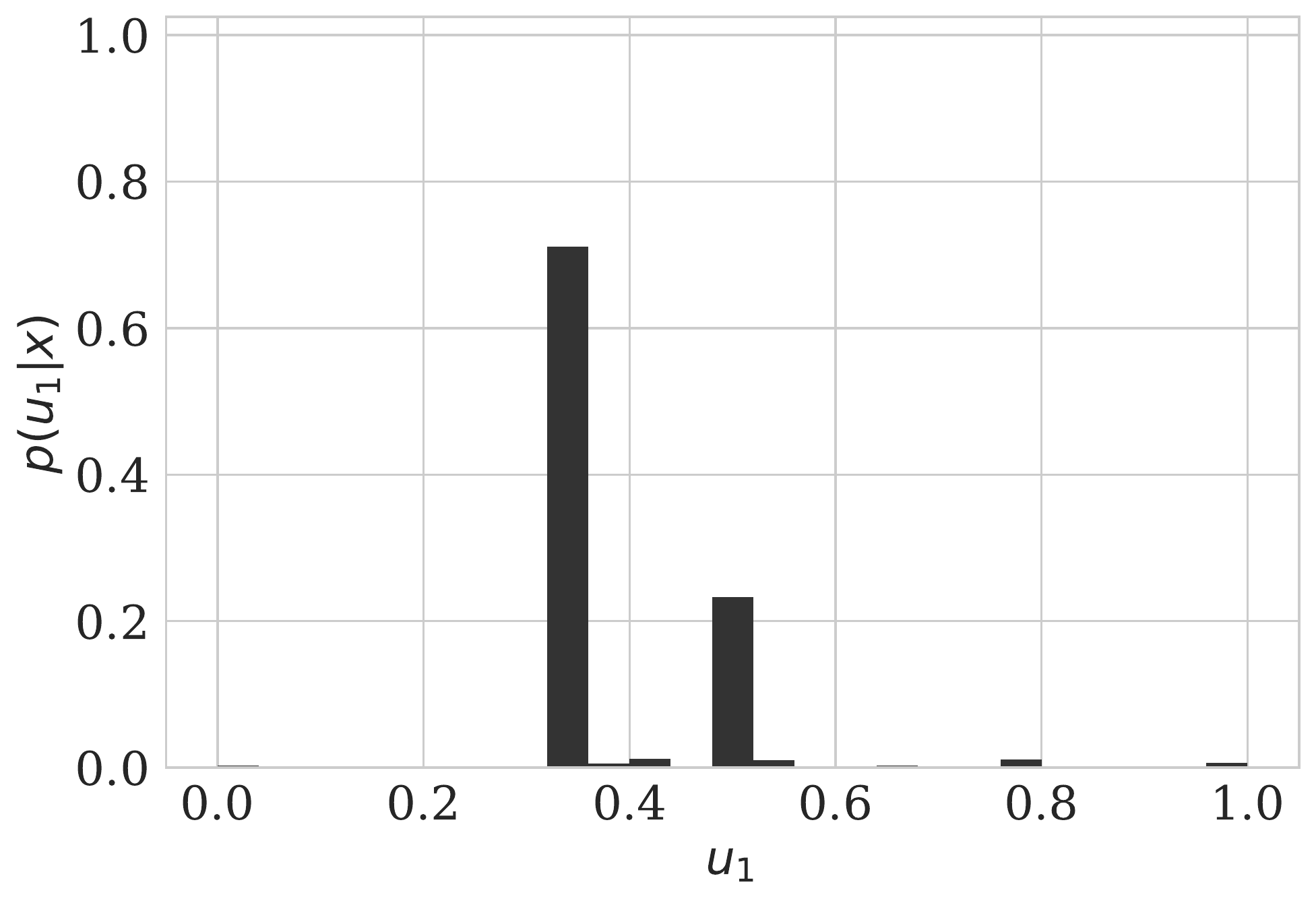}\\
    \qquad \quad$p_{closure}=0$ \quad\quad \qquad \qquad \qquad \qquad \quad $p_{closure}=0.2$ \\
    \vspace{1cm}
        \includegraphics[width=0.45\linewidth]{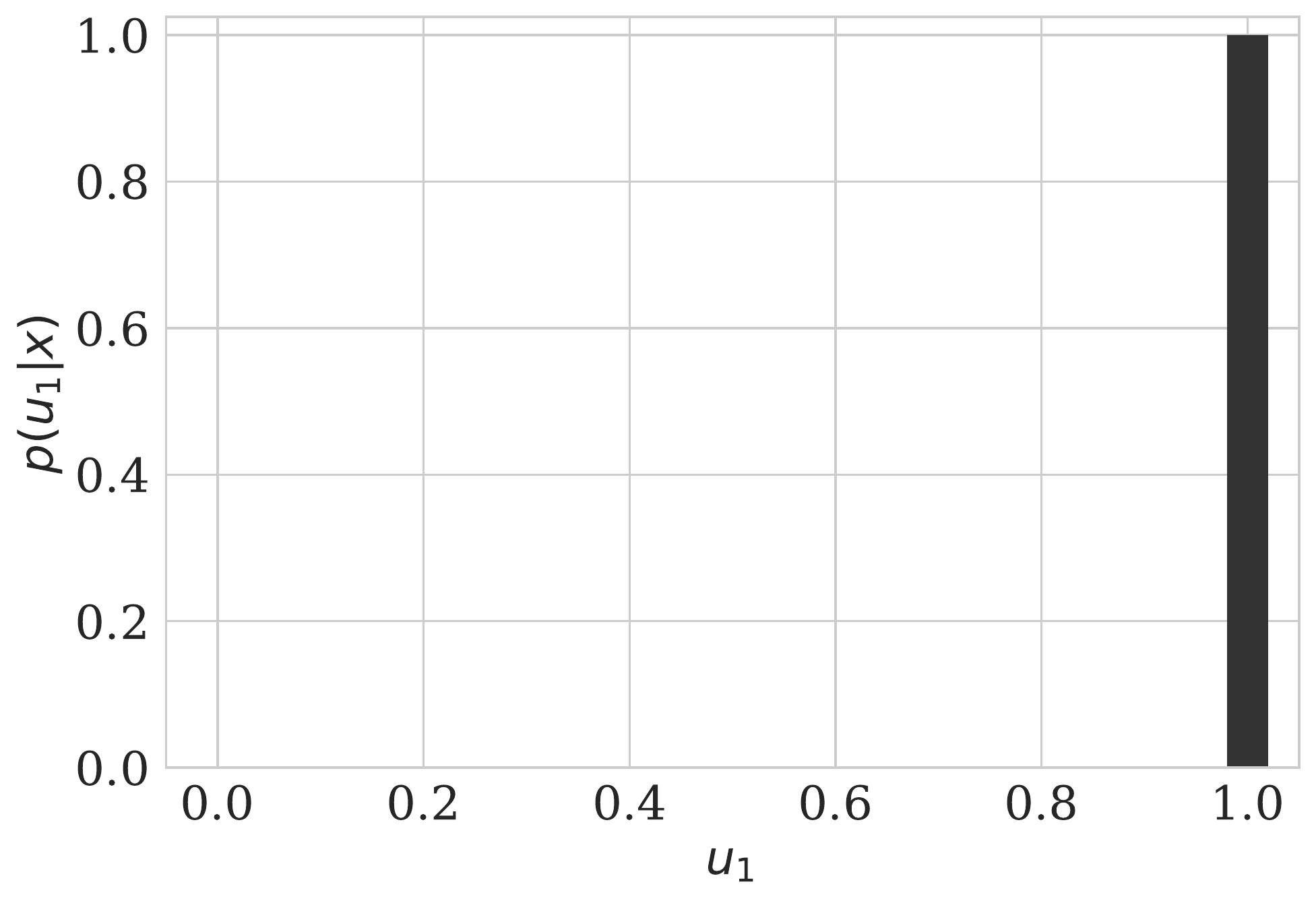}
    \includegraphics[width=0.45\linewidth]{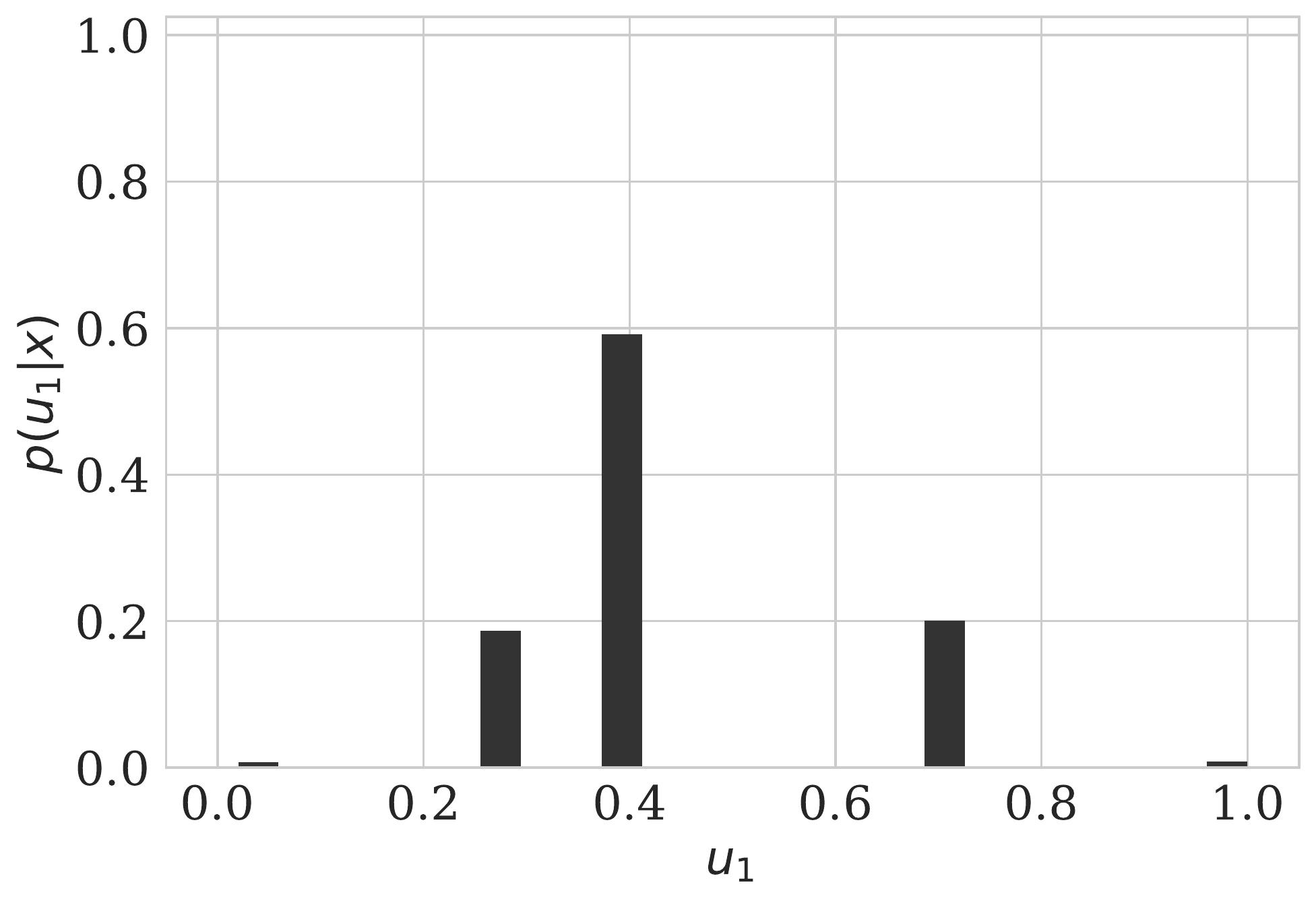}\\
     \qquad \quad Always\quad\quad \qquad \qquad \qquad \qquad \qquad\quad 5-periodic \\
    \caption{Histograms of the firm's level of tax-evasion throughout its lifetime (100 episodes, 250 steps per episode).}
    \label{fig:hist_u1}
\end{figure}
Fig.~\ref{fig:hist_u1} shows the frequency histograms of the firm's optimal level of tax-evasion over 25000 state-decision pair samples (obtained from our trained DQN, over 100 episodes where the firm was allowed to evolve for 250 steps, as previously mentioned).  
We observe that there is no variability in the case where closure is always available (the firm always uses the closure option and conceals as much profit as possible). In the cases where the option is offered stochastically or periodically there is more significant variability in the optimal $[u]_1$ (Fig.~\ref{fig:hist_u1}, top row, and bottom-right), although we observe that the set of values for $[u]_1$ used by the DQN is sparse. 

To gain insight into just how the values observed in the histograms depend on the firm's state, we used decision-tree classifiers. 
Fitting a decision tree to the outputs of the network is a commonly-used approach for discovering patterns in the learned policy. We opted for a shallow decision tree (depth = 3) to the same 25000 outputs $[u]_1$, with a high threshold for splitting ($10^{-4}$). We kept the tree classifier ``naive'' in order to be able to gain high-level intuition on the decision policy's structure. 

\begin{figure}[!htbp]
    \centering
    \includegraphics[width=0.45\linewidth]{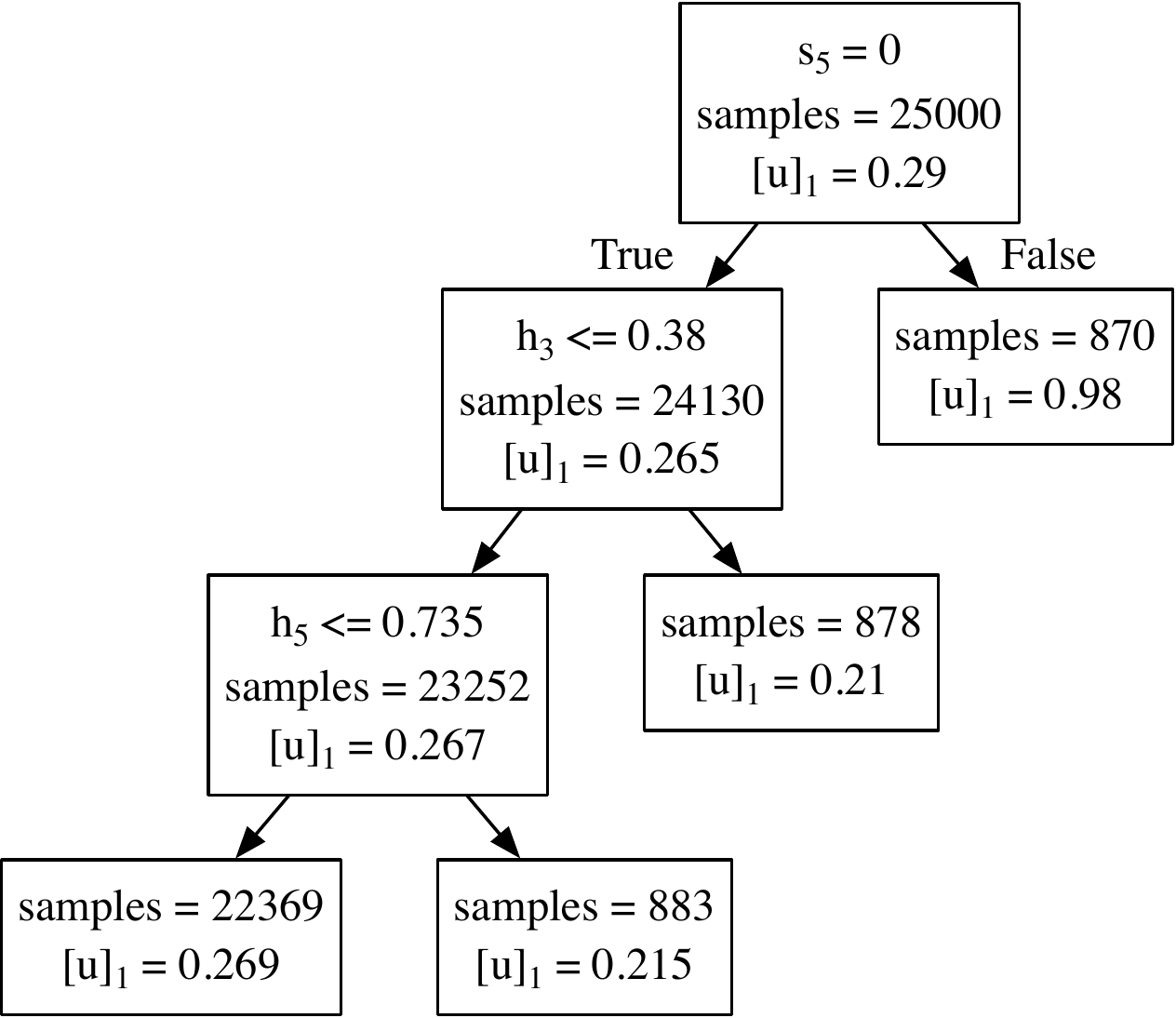}
    \includegraphics[width=0.45\linewidth]{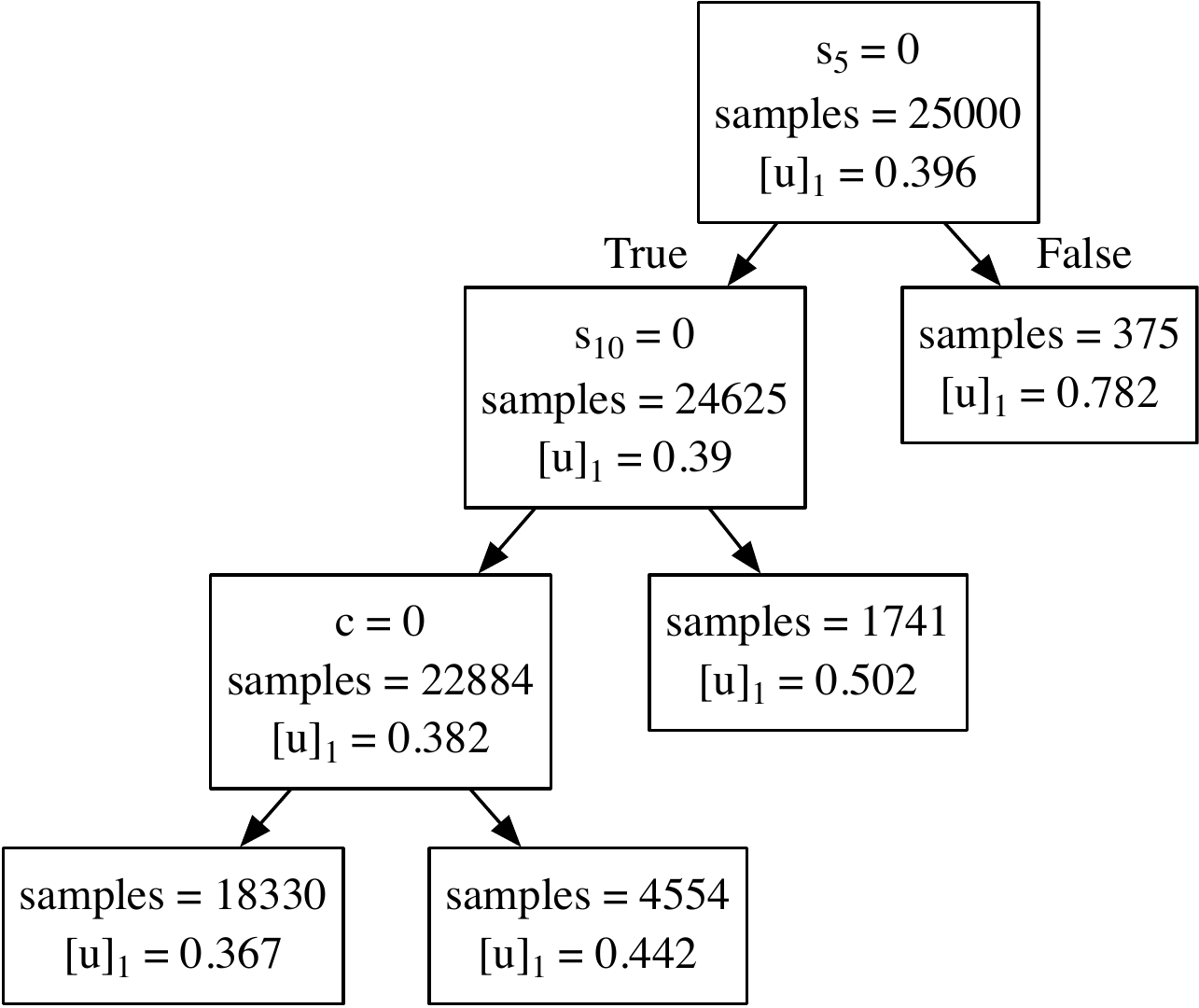}\\
    $p_{closure}=0$ \quad \quad \qquad \qquad \qquad \qquad \qquad $p_{closure}=0.2$
    \caption{Decision trees analyzing the firm's levels of tax-evasion $[u]_1$ (left: closure never available, right: closure available with a 20\% probability) using samples from 100 episodes of 250 steps each. Data were fit to purposely shallow (depth-3) decision trees with a high threshold for splitting, in order to glean information on the high-level structure of the policy. }
    \label{fig:tree_u1}
\end{figure}

Fig.~\ref{fig:tree_u1} illustrates the trees obtained for the  cases of $p_{closure} = 0$ and $0.2$. In the tree nodes, the binary $s_i$ stand for the firm's tax status in terms of the i-th element of {\cal S} (see state space description following Eq.~\ref{eq:firm_dynamics}), e.g., $s_5=0$ means that the firm's tax status is {\em not} the fifth element of $\cal S$, so that the firm is {\em not} being audited for its last five tax years; $h_i$ denotes the i-th element of the firm's tax history vector $h$, i.e., the amount of profit it concealed $5-i+1$ years ago; $c\in \{0,1\}$ denotes whether closure is available to the firm or not; and $samples$ denotes the number of samples (out of the 25000 total) to which each case applied.
The decision tree for $p_{closure=0}$ (Fig.~\ref{fig:tree_u1}-left) indicates that if closure is never offered, the firm opts for very high tax evasion ($[u]_1=0.98$, top-right leaf of the tree) only immediately after audits that ``cover'' the last 5 years. The remainder of the time, the firm 
almost always conceals 27\% of its profit (bottom-left leaf of the decision tree), and any deviations from that value depend mainly on its history $h$ of tax evasion (e.g., whether 3 years ago it concealed more or less than 38\% of its profit - see left ``child'' of the tree's root node). 

When $p_{closure}=0.2$ (Fig.~\ref{fig:tree_u1}-right), the firm again uses a high $[u]_1 = 0.78$ immediately after (rare) audits; for the majority of its time it uses two tax evasion levels, $[u]_1=0.44$ or $[u]_1=0.37$ depending on whether closure is ($c=1$) or is not ($c=0$) available, respectively.
For the 5-periodic closure scenario, the classifier (not shown) indicated that when the firm is 3 or 4 years away from the next closure, it uses a near-average $[u]_1 \approx 0.38$. If the closure option is less than 3 years away, and the firm has been recently audited ($<5$ years ago), then its tax-evasion goes up to $[u]_1 \approx 0.67$.

To glean additional information on the structure of the DQN policy, we looked for patterns in the tax evasion decisions based on i) the tax status of the firm (i.e., whether it is being audited, using the closure option, or left to evolve with 1-5 years since its last audit or closure, as detailed in Sec.~\ref{sec:model_eq}), and ii) the cumulative tax evasion ``stored'' in the firm's history ($h_k$) within the 5-year statute of limitations, this representing a kind of ``amount at risk'' that the firm would be liable for if it were to be audited.   
\begin{figure}[htbp]
    \centering
    \includegraphics[width=0.49\linewidth]{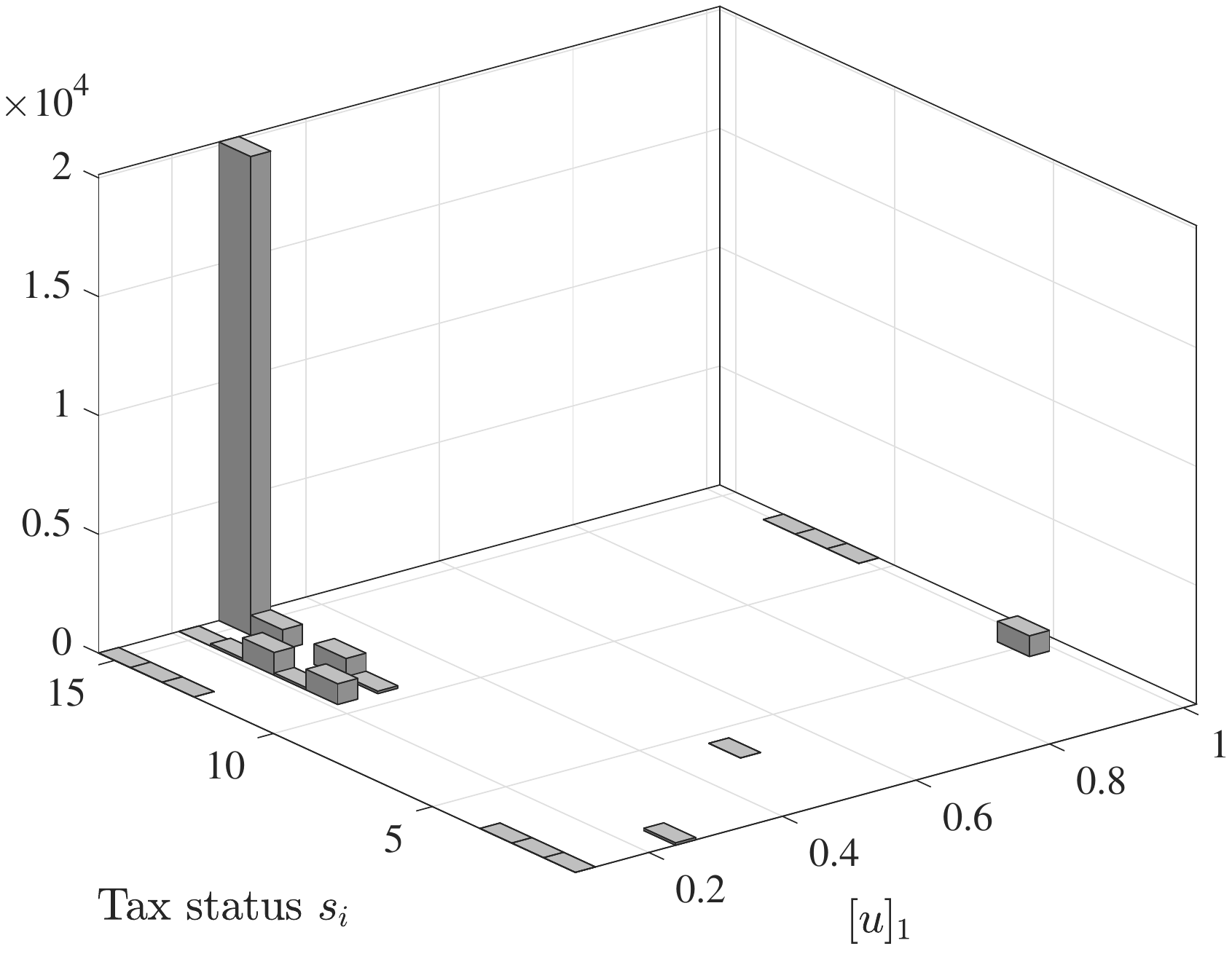}
    \includegraphics[width=0.49\linewidth]{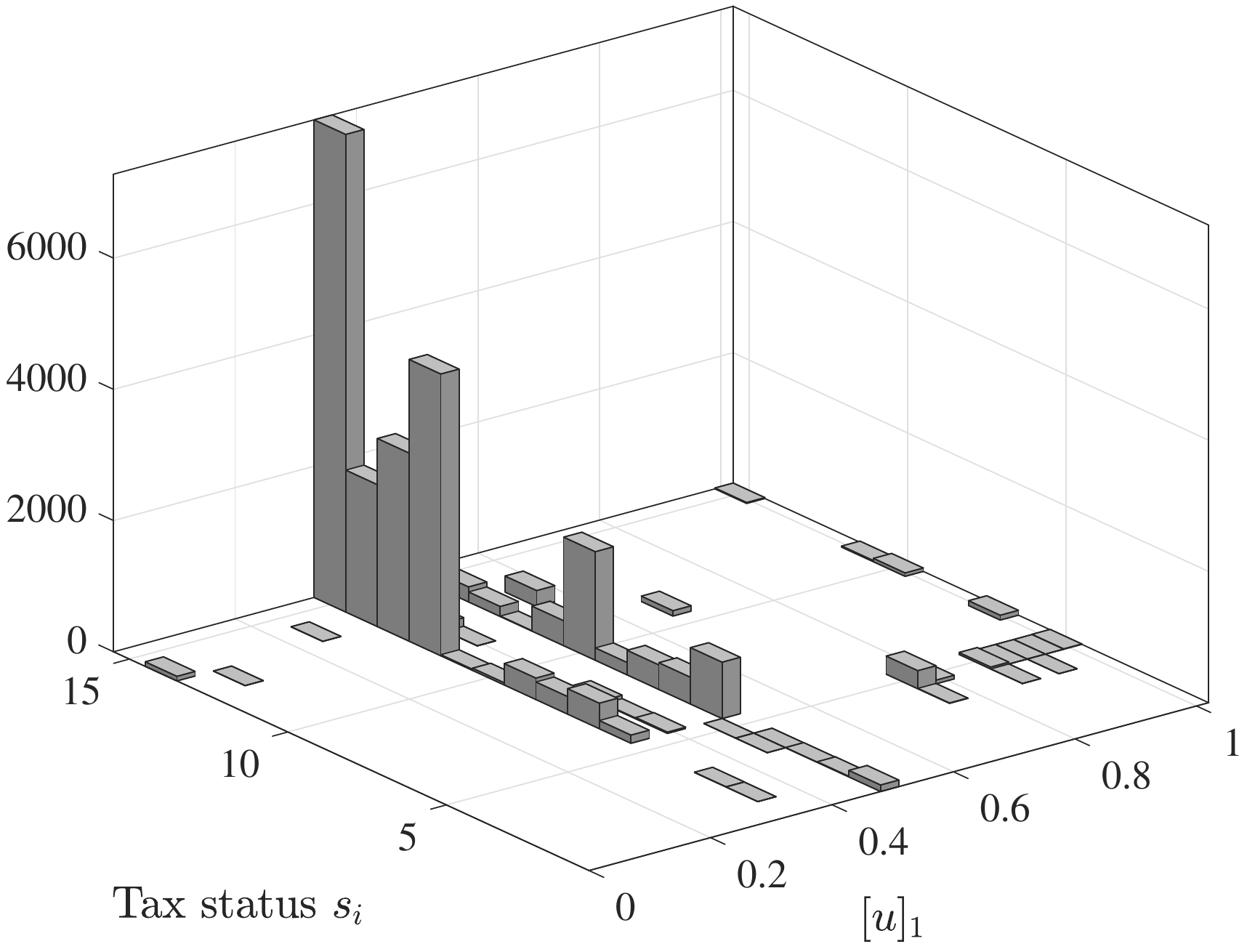}\\
    \quad \qquad $p_{closure}=0$ \quad\quad ~~ \qquad \qquad \qquad \qquad \qquad $p_{closure}=0.2$ \\
    \caption{Histograms showing the distribution of 25000 tax evasion decision samples ($[u]_1$) from 100 episodes of 250 steps each. Axes on the horizontal plane correspond to the level of tax evasion (0 to 1) and the integer-coded tax status of the firm (1-15, as explained in Sec.~\ref{sec:model_eq}).}
    \label{fig:hist_3d_15}
\end{figure}

Fig.~\ref{fig:hist_3d_15} shows histograms of the firm's decisions according to level  of tax evasion ($[u]_1$) and the tax status of the firm (shown as an integer between 1 and 15 representing states in $\cal{S}$, as per Sec.~\ref{sec:model_eq}). In the left histogram, where closure is never available, we observe that the firm spends most of its time in the tax status 15 (which corresponds to the firm having been unaudited for 5 or more years) and its level of tax evasion is near 0.28 (this matches the decision tree analysis above). Also noteworthy is the fact that the firm consistently uses $[u]_1=1$ when its tax status is 5  
(the firm being audited for its last 5 tax filings). 

In the right histogram of Fig.~\ref{fig:hist_3d_15}, the closure option is 
available with probability $0.2$, and if we were to sum over the tax status axis 
we would obtain the top-right histogram of Fig.~\ref{fig:hist_u1}. The firm 
generally uses a higher level of tax evasion ($[u]_1\approx0.35-0.5$). The 
broader spread of the samples over the tax status axis compared with the 
previous case (closure never available) indicates that the firm uses the option 
when it can, thereby ``erasing'' any tax evasion history and thus finds itself 
more often in a tax status of 5-10 (corresponding to closure 
being used for the last 1-5 tax filings of the firm) or 11-15 (the firm being unaudited for 1-5 years in the past).

\begin{figure}[htbp]
    \centering
    \includegraphics[width=0.49\linewidth]{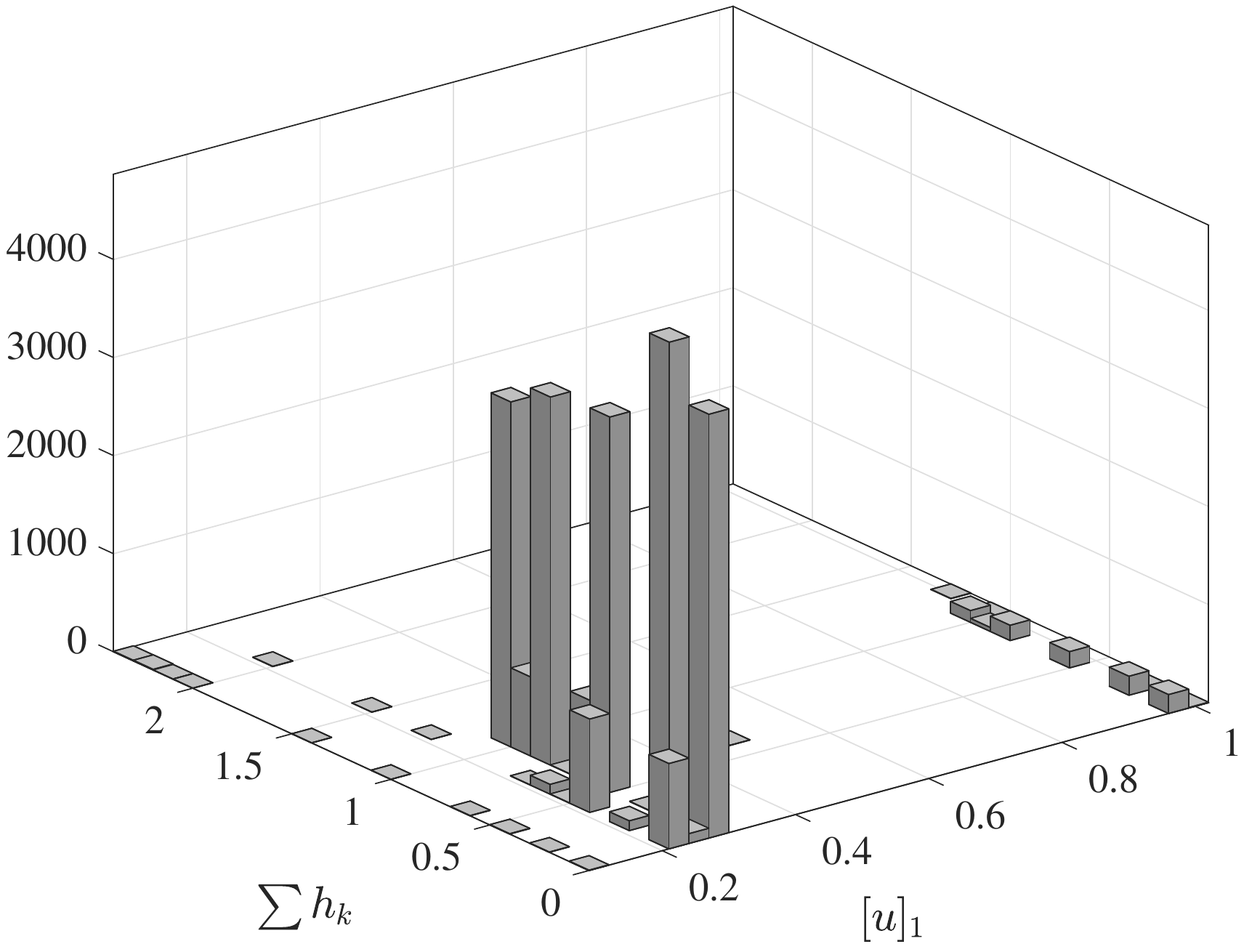}
    \includegraphics[width=0.49\linewidth]{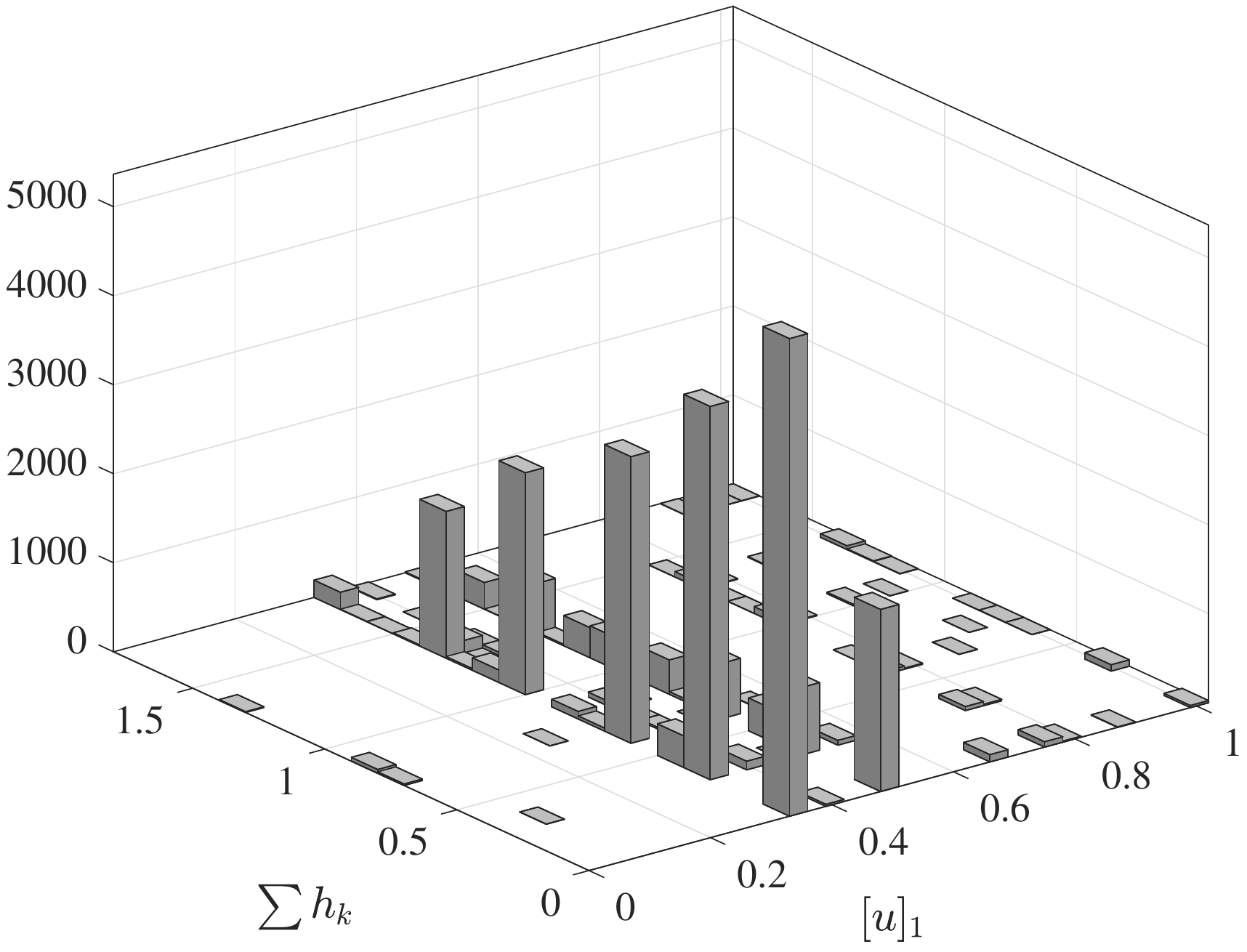}\\
     $p_{closure}=0$ \quad\quad \qquad \qquad \qquad \qquad \qquad $p_{closure}=0.2$ \\
    \caption{Histograms showing the distribution of 25000 tax evasion decision samples ($[u]_1$) from 100 episodes of 250 steps each. Axes on the horizontal plane correspond to the level of tax evasion (0 to 1) and the sum of the past history of tax evasion decisions, as explained in Sec.~\ref{sec:model_eq}).}
    \label{fig:hist_3d_sumh}
\end{figure}

Besides grouping the firm's decisions by tax status, we examined how the firm behaves based on the part of its state, $h_k$, which contains its past tax evasion decisions (up to five) which are still within the statute of limitations (see Sec.~\ref{sec:model_eq}). Because we have quantized $[u]_1$ in steps of $0.01$, and because of the structure of $h_k$ as the firm evolves via Eq.~\ref{eq:firm_dynamics}, it is difficult to visualize the firm's policy over that entire set. It is instructive, however, to consider the sum of the elements of $h_k$ (which is proportional to the total amount the firm has failed to disclose) as a proxy variable for the amount at risk if it the firm were to be audited, and examine how it affects tax evasion by the firm. We expect that a ``good'' policy would reduce tax evasion ($[u]_1$) when that sum increases, which is precisely what happens. Fig.~\ref{fig:hist_3d_sumh} shows the histograms of the firm's level of tax evasion and sum of its past decisions (up to five or up to the last time it was audited or used the closure option, whichever is smaller). In the left histogram, where closure is never available, we observe that although the firm conceals approximately 30\% of its profit most of the time, it sometimes decides to be completely dishonest with $[u]_1$ at 1, when the amount it is potentially ``on the hook'' for ($\sum h_k$) is small  (between 0 and 1.2) but becomes more honest (with $[u]_1$ at 0 or 0.2) when that amount is larger.

In the right histogram of Fig.~\ref{fig:hist_3d_sumh}, the closure option is available with probability 0.2, and there are occurrences of $[u]_1=1$ throughout the range of values for $\sum h_k$. This is explained by the fact that the usage of closure allows the firm to ``wipe the slate clean'' so that it is less deterred by the fact that it has accumulated a history of tax evasion. The downward trend present in the bars near $[u]_1=0.35$ is because by using the closure option whenever possible (thereby zeroing out $h_k$), it is more likely for the firm to find itself with a lower value of  $\sum h_k$.

\subsection{Tax policy implications}
\label{sec:policy_implications}
With an eye towards making policy recommendations for the ``canonical'' firm   ($\lambda=2.6$) we observe that, based on the results of Sec.~\ref{sec:risk-averse-results}, the more frequently the closure option is offered by the government, 
the higher the firm's expected utility (see left column in Tab. ~\ref{tab:compare_constant}), and - correspondingly - the lower the amount of tax-revenue collected. Thus, it appears that government should avoid using this type of tax amnesty because it encourages tax evasion, and instead reinforce auditing mechanisms. 

Also, the analysis of the DQN policy in Sec.~\ref{sec:lam26} suggests ways in which the tax authority could re-allocate auditing resources towards firms which are in  states associated with the highest tax evasion.  In particular, under the current regime, most auditing resources are devoted to firms which have not been audited for five years and thus have past tax filings which are about to pass beyond the statute of limitations. The histograms and decision tree analysis of the firm's policy shows that tax evasion is high immediately after an audit, suggesting that the audit probabilities should be distributed more ``evenly'' on $\cal S$, to improve the chance of catching tax evaders that were audited just one year ago. 

Finally, Fig.~\ref{fig:k1} gives guidance for the expected reduction in tax evasion  as the firm's risk aversion increases. Of course, it is not easy to directly influence firm's attitudes to make them more risk averse. However, the relationship between average $[u]_1$ and $\lambda$ provides an opportunity for optimizing the allocation of auditing resources among various categories of firms (grouped, for example, by size or sector of economic activity), with fewer audits for the very risk-averse, more for those who are less so, once each group's risk aversion coefficients are estimated (this can be done empirically by examining  tax audits to measure tax evasion within each group, and estimating that group's $\lambda$ as we did in Sec.~\ref{sec:estimates_risk}).

\section{Conclusions}
\label{sec:conclusions}

This work is part of a research program whose aim is to provide governments with quantitative tools which can be used to combat tax evasion and guide tax policy. A prerequisite for the design of effective policies is to be able to understand, in quantitative terms, the behavior of tax evaders. 
Towards that end, we addressed the problem of determining the behavior expected of a self-interested risk-averse firm which aims to maximize its long-term revenues, in a tax system whose features include tax rates, random audits, penalties for tax evasion and occasional tax amnesties. The practical importance of the problem is significant: solving it allows one to estimate tax revenues, to identify measures and parameter values that make self-interested entities behave more honestly, and to gauge the effectiveness of current or planned tax policies. 

The dynamics of the firm's (stochastic) evolution, combined with the rules of the tax system and the nonlinearity of the firm's reward function (owing to the fact that the firm is generally risk-averse), give rise to a stochastic optimal decision problem in which the associated Bellman equation is difficult to solve using exact methods. To address that challenge, we made use of recent developments in function approximation and neural networks and constructed a Deep Q-learning Network (DQN) which ``learns'' the optimal firm policy. The neural network was trained to ``store'' the firm's optimal long-term revenues, given a starting state and decision. DQN was used to efficiently ``learn'' the optimal firm decisions through simulations of the firm's state evolution. 

The DQN approach was first validated by setting our model to the special case of risk neutrality and comparing the results thus obtained (optimal policy and long-term firm revenues) to the exact solution computed via DP \citep{goumagias2012decision}. We subsequently demonstrated that we can compute the firm's optimal policy and corresponding tax revenues for the government in the ``full'' model which includes both risk-aversion (i.e., non-linearity in the reward function) and the tax amnesty (``closure'') option. 
We note that, in our particular case, Deep Learning was successful in approximating the firm's reward function and finding its optimal decisions where other approximation methods failed to converge (we experimented extensively with Approximate Dynamic Programming, various implementations of Q-learning and SARSA algorithms, and neural network architectures which served as function approximators). 

One of the contributions made possible by our approach is that it can be used to infer the risk aversion coefficient of a typical taxpayer from empirical data, and thus subsequently evaluate that taxpayer's reactions under various scenarios of tax amnesty availability, or other parameter change (i.e., increase in the audit rates or penalties). Using Greece as a case study, 
we estimated the risk aversion coefficient of the average  firm to be approximately $\lambda = 2.6$, based on empirical evidence that puts the level of the Greek ``hidden'' economy at approximately 40\% \citep{artavanis2016measuring}. We also compared tax revenues for a series of policies used there; our results provide evidence against the use of tax amnesties as tax revenue collection tools, even within economies with persistent and endemic tax evasion, as we there is a negative relationship between the predictability (or indeed existence) of tax amnesties and tax revenue.
Although we have used Greece as a case study here, in part for the sake of concreteness, the proposed approach is adaptable to different taxation schemes and can easily be ``tuned'' to reflect the values of various tax-parameters, such as audit rates, which are known to the government. 

Opportunities for further work include the use of the very recent sample-efficient actor-critic algorithm with experience replay \citep{wang2016sample}, which could enable stable learning in continuous action spaces (without having to discretize the firm's decisions); efficient reward scaling, to handle reward values across many orders of magnitude similarly to \citet{van2016learning}; and the use of Recurrent Q-Learning to possibly reduce some state features, e.g., the firm's behavior in the past five-year window.

An interesting (and massive) computational study  which has now been made possible in light of the present work, involves recording the effects of altering the various tax parameters on the behavior of the firm, so that one could compute the ``degree of honesty'' of the firm as a function of the parameters, in the spirit of the maps given in \citet{goumagias2012decision}. 

Finally, we also envision extensions of this work with learning models that generalize over different values of the tax rate $r$ or the risk aversion coefficient $\lambda$ (instead of having to be trained separately for particular values), or that also optimize selected model parameters simultaneously with the firm's decisions. Although some parameters, such as $\lambda$, are generally considered exogenous in forming the firms' risk preferences, optimizing others, especially the tax rate and penalty factor would be of particular interest for the purposes of maximizing tax revenue.

\section{References}
\bibliographystyle{apa}
\bibliography{refs}

\newpage
\appendix
\section{State dynamics}
\label{ap:statedynamics}

The parameters of the basic state equation \ref{eq:firm_dynamics} are 
as in \cite{goumagias2012decision} but are also given here for the purposes of review:

\begin{equation}
    \label{eq:ahbn}
    A = \begin{bmatrix}
    0 & & \\
     & 0 & \\
     & & H \\
    \end{bmatrix},\ H = \begin{bmatrix}
    0 & 1 & 0 & 0 & 0 \\
    0 & 0 & 1 & 0 & 0 \\
    0 & 0 & 0 & 1 & 0 \\
    0 & 0 & 0 & 0 & 1 \\
    0 & 0 & 0 & 0 & 0 \\
    \end{bmatrix},\ B = \left\lbrack \begin{pmatrix}
    0 & 0 \\
     \vdots & \vdots \\
    0 & 0 \\
    0 & 1 \\
    \end{pmatrix} \right\rbrack,\ n_{k} = \ \begin{bmatrix}
    \omega_{k} \\
    \epsilon_{k} \\
    0_{5 \times 1} \\
    \end{bmatrix}.    
\end{equation}

The scalar $\epsilon_k$ corresponds to the government determining whether to offer the closure option; this occurs with some fixed probability, $p_o$, each year so that:
\begin{equation}
     \Pr\left( \epsilon_{k} = 1 \right) = \ \left\{ \begin{matrix}
     p_{o} \\
     1 - p_{o} \\
     \end{matrix} \right.\ \text{\ \ \ \ }\begin{matrix}
     if\ i = 1 \\
     if\ i = 2 \\
     \end{matrix}\text{\ \ \ \ \ }\begin{matrix}
     \left( \text{option\ available} \right) \\
     \left( \text{option\ not\ available} \right). \\
     \end{matrix}
     \label{eq:po_year}
 \end{equation}

The scalar $\omega_k \in \{1,...,|\emph{S}|\}$ is a random variable corresponding to the transitions that the firm undergoes in \emph{S} (e.g., tax audits) depend on its current state and its decision $\left\lbrack u_{k} \right\rbrack_{2}$ to accept or reject the closure option (if offered):
 \begin{multline}
     \Pr\left( \omega_{\kappa} = i | x_{k} = \lbrack j,\ q,\ h_{k}^{T} \rbrack^{T}, \lbrack u_{k} \rbrack_{2} = m \right) \\ = P_{\text{qij}}\left( m \right),
     i,j \in \left\{1, \ldots, 15 \right\},\ q \in \{1,2\}
     \label{eq:po_year_random}    
 \end{multline}
where we use $1,...,15$ as labels for states in $\emph{S}$. For $q$ and $u$ fixed, $P_{\text{qij}}\left( m \right)$ forms a Markov matrix governing the firm's transitions in $S$:
\begin{equation}
     P_{\text{qij}}\left( m \right) = \left\{ \begin{matrix}
     \left\lbrack M_{\text{no}} \right\rbrack_{\text{ij}} \\
     \left\lbrack M_{a} \right\rbrack_{\text{ij}} \\
     \left\lbrack M_{d} \right\rbrack_{\text{ij}} \\
     \end{matrix} \right.\ \text{\ \ \ }\begin{matrix}
     if\ q = 2,\ \ \ \ \forall\ m \\
     if\ q = 1,\ \ \ \ m = 1 \\
     if\ q = 1,\ \ \ \ m = 2 \\
     \end{matrix}\text{\ \ \ \ }\begin{matrix}
     \text{(no\ option)} \\
     \text{(option\ taken)} \\
     \text{(option\ declined)} \\
     \end{matrix}    
 \end{equation}
 and the $M_{\text{no}}, M_{a}$, and $M_{d}$ are as in \cite{goumagias2012decision} but are also given in
 Appendix~B for the purposes of review.

\section{Markov transition matrices}
\label{sec:appendixB}
\begin{table}[htbp]
\vspace*{-1cm}
\centering
	\begin{multline*}
		M_{\text{no}} = \\ \left[\begin{smallmatrix}
	0.0025 & 0.0025 & 0.0025 & 0.0025 & 0.0025 & 0 & 0 & 0 & 0 & 0 & 0 & 0 & 0 & 0 & 0 \\
	0 & 0 & 0 & 0 & 0 & 0.0025 & 0.0025 & 0.0025 & 0.0025 & 0.0025 & 0.0025 & 0 & 0 & 0 & 0 \\
	0 & 0 & 0 & 0 & 0 & 0 & 0 & 0 & 0 & 0 & 0 & 0.0025 & 0 & 0 & 0 \\
	0 & 0 & 0 & 0 & 0 & 0 & 0 & 0 & 0 & 0 & 0 & 0 & 0.0025 & 0 & 0 \\
	0 & 0 & 0 & 0 & 0 & 0 & 0 & 0 & 0 & 0 & 0 & 0 & 0 & 0.04 & 0.04 \\
	0 & 0 & 0 & 0 & 0 & 0 & 0 & 0 & 0 & 0 & 0 & 0 & 0 & 0 & 0 \\
	0 & 0 & 0 & 0 & 0 & 0 & 0 & 0 & 0 & 0 & 0 & 0 & 0 & 0 & 0 \\
	0 & 0 & 0 & 0 & 0 & 0 & 0 & 0 & 0 & 0 & 0 & 0 & 0 & 0 & 0 \\
	0 & 0 & 0 & 0 & 0 & 0 & 0 & 0 & 0 & 0 & 0 & 0 & 0 & 0 & 0 \\
	0 & 0 & 0 & 0 & 0 & 0 & 0 & 0 & 0 & 0 & 0 & 0 & 0 & 0 & 0 \\
	0.9975 & 0.9975 & 0.9975 & 0.9975 & 0.9975 & 0 & 0 & 0 & 0 & 0 & 0 & 0 & 0 & 0 & 0 \\
	0 & 0 & 0 & 0 & 0 & 0.9975 & 0.9975 & 0.9975 & 0.9975 & 0.9975 & 0.9975 & 0 & 0 & 0 & 0 \\
	0 & 0 & 0 & 0 & 0 & 0 & 0 & 0 & 0 & 0 & 0 & 0.9975 & 0 & 0 & 0 \\
	0 & 0 & 0 & 0 & 0 & 0 & 0 & 0 & 0 & 0 & 0 & 0 & 0.9975 & 0 & 0 \\
	0 & 0 & 0 & 0 & 0 & 0 & 0 & 0 & 0 & 0 & 0 & 0 & 0 & 0.96 & 0.96 \\
	\end{smallmatrix}\right]
	\end{multline*}
	\caption{Transition Probabilities $M_{\text{no}}$ : Closure is not available}
\end{table}
\vspace{-0.5cm}
\begin{table}[htbp]
	\centering
	\begin{equation*}
M_{a} = \left[\begin{smallmatrix}
0 & 0 & 0 & 0 & 0 & 0 & 0 & 0 & 0 & 0 & 0 & 0 & 0 & 0 & 0 \\
0 & 0 & 0 & 0 & 0 & 0 & 0 & 0 & 0 & 0 & 0 & 0 & 0 & 0 & 0 \\
0 & 0 & 0 & 0 & 0 & 0 & 0 & 0 & 0 & 0 & 0 & 0 & 0 & 0 & 0 \\
0 & 0 & 0 & 0 & 0 & 0 & 0 & 0 & 0 & 0 & 0 & 0 & 0 & 0 & 0 \\
0 & 0 & 0 & 0 & 0 & 0 & 0 & 0 & 0 & 0 & 0 & 0 & 0 & 0 & 0 \\
1 & 1 & 1 & 1 & 1 & 1 & 1 & 1 & 1 & 1 & 1 & 0 & 0 & 0 & 0 \\
0 & 0 & 0 & 0 & 0 & 0 & 0 & 0 & 0 & 0 & 0 & 1 & 0 & 0 & 0 \\
0 & 0 & 0 & 0 & 0 & 0 & 0 & 0 & 0 & 0 & 0 & 0 & 1 & 0 & 0 \\
0 & 0 & 0 & 0 & 0 & 0 & 0 & 0 & 0 & 0 & 0 & 0 & 0 & 1 & 0 \\
0 & 0 & 0 & 0 & 0 & 0 & 0 & 0 & 0 & 0 & 0 & 0 & 0 & 0 & 1 \\
0 & 0 & 0 & 0 & 0 & 0 & 0 & 0 & 0 & 0 & 0 & 0 & 0 & 0 & 0 \\
0 & 0 & 0 & 0 & 0 & 0 & 0 & 0 & 0 & 0 & 0 & 0 & 0 & 0 & 0 \\
0 & 0 & 0 & 0 & 0 & 0 & 0 & 0 & 0 & 0 & 0 & 0 & 0 & 0 & 0 \\
0 & 0 & 0 & 0 & 0 & 0 & 0 & 0 & 0 & 0 & 0 & 0 & 0 & 0 & 0 \\
0 & 0 & 0 & 0 & 0 & 0 & 0 & 0 & 0 & 0 & 0 & 0 & 0 & 0 & 0
	\end{smallmatrix}\right]
	\end{equation*}
	\caption{Transition Probabilities $M_{a}$ : Closure is available and the firm decides to use it.}
\end{table}
\begin{table}[htbp]
\vspace*{-2.2cm}
	\centering
	\begin{multline*}
M_{d} =\\ \left[\begin{smallmatrix}
0.0075 & 0.0075 & 0.0075 & 0.0075 & 0.0075 & 0 & 0 & 0 & 0 & 0 & 0 & 0 & 0 & 0 & 0 \\
0 & 0 & 0 & 0 & 0 & 0.0075 & 0.0075 & 0.0075 & 0.0075 & 0.0075 & 0.0075 & 0 & 0 & 0 & 0 \\
0 & 0 & 0 & 0 & 0 & 0 & 0 & 0 & 0 & 0 & 0 & 0.0075 & 0 & 0 & 0 \\
0 & 0 & 0 & 0 & 0 & 0 & 0 & 0 & 0 & 0 & 0 & 0 & 0.0075 & 0 & 0 \\
0 & 0 & 0 & 0 & 0 & 0 & 0 & 0 & 0 & 0 & 0 & 0 & 0 & 0.12 & 0.12 \\
0 & 0 & 0 & 0 & 0 & 0 & 0 & 0 & 0 & 0 & 0 & 0 & 0 & 0 & 0 \\
0 & 0 & 0 & 0 & 0 & 0 & 0 & 0 & 0 & 0 & 0 & 0 & 0 & 0 & 0 \\
0 & 0 & 0 & 0 & 0 & 0 & 0 & 0 & 0 & 0 & 0 & 0 & 0 & 0 & 0 \\
0 & 0 & 0 & 0 & 0 & 0 & 0 & 0 & 0 & 0 & 0 & 0 & 0 & 0 & 0 \\
0 & 0 & 0 & 0 & 0 & 0 & 0 & 0 & 0 & 0 & 0 & 0 & 0 & 0 & 0 \\
0.9925 & 0.9925 & 0.9925 & 0.9925 & 0.9925 & 0 & 0 & 0 & 0 & 0 & 0 & 0 & 0 & 0 & 0 \\
0 & 0 & 0 & 0 & 0 & 0.9925 & 0.9925 & 0.9925 & 0.9925 & 0.9925 & 0.9925 & 0 & 0 & 0 & 0 \\
0 & 0 & 0 & 0 & 0 & 0 & 0 & 0 & 0 & 0 & 0 & 0.9925 & 0 & 0 & 0 \\
0 & 0 & 0 & 0 & 0 & 0 & 0 & 0 & 0 & 0 & 0 & 0 & 0.9925 & 0 & 0 \\
0 & 0 & 0 & 0 & 0 & 0 & 0 & 0 & 0 & 0 & 0 & 0 & 0 & 0.88 & 0.88
	\end{smallmatrix}\right]
	\end{multline*}
	\caption{Transition Probabilities \(M_{d}\) : Closure is available and the firm decides not to use it.}
\end{table} 

\end{document}